\documentclass{article}

\usepackage{arxiv}

\usepackage[utf8]{inputenc} 
\usepackage[T1]{fontenc}    
\usepackage{hyperref}       
\usepackage{url}            
\usepackage{booktabs}       
\usepackage{amsfonts}       
\usepackage{nicefrac}       
\usepackage{microtype}      
\usepackage{graphicx}
\usepackage{doi}

\usepackage{multirow}
\usepackage{subfig}
\usepackage{float}

\title{A Large Visual, Qualitative and Quantitative
Dataset of Web Pages}

\author{ \href{https://orcid.org/0000-0001-6715-191X}{\includegraphics[scale=0.06]{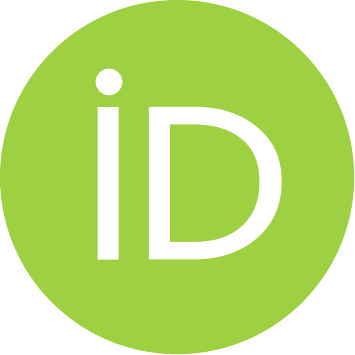}\hspace{1mm}Christian Mejia-Escobar}

\\
	Central University of Ecuador\\
	P.O. Box 17-03-100\\
	Quito, Ecuador\\
	\texttt{cimejia@uce.edu.ec} \\
	\And
	\href{https://orcid.org/0000-0001-6805-3633}{\includegraphics[scale=0.06]{orcid.pdf}\hspace{1mm}Miguel Cazorla} \\
	Institute for Computer Research\\
	University of Alicante\\
	P.O. Box 99. 03080, Spain \\
	\texttt{miguel.cazorla@ua.es} \\
	\And
	\href{https://orcid.org/0000-0003-4495-6912}{\includegraphics[scale=0.06]{orcid.pdf}\hspace{1mm}Ester Martinez-Martin} \\
	Institute for Computer Research\\
	University of Alicante\\
	P.O. Box 99. 03080, Spain \\
	\texttt{ester@gcloud.ua.es} \\
	
}

\hypersetup{
pdftitle={A template for the arxiv style},
pdfsubject={q-bio.NC, q-bio.QM},
pdfauthor={David S.~Hippocampus, Elias D.~Striatum},
pdfkeywords={First keyword, Second keyword, More},
}

\begin{document}
\maketitle

\begin{abstract}
The \textit{World Wide Web} is not only one of the most important platforms of communication and information at present, but also an area of growing interest for scientific research. This motivates a lot of work and projects that require large amounts of data. However, there is no dataset that integrates the parameters and visual appearance of Web pages, because its collection is a costly task in terms of time and effort. With the support of various computer tools and programming scripts, we have created a large dataset of 49,438 Web pages. It consists of visual, textual and numerical data types, includes all countries worldwide, and considers a broad range of topics such as art, entertainment, economy, business, education, government, news, media, science, and environment, covering different cultural characteristics and varied design preferences. In this paper, we describe the process of collecting, debugging and publishing the final product, which is freely available. To demonstrate the usefulness of our dataset, we expose a binary classification model for detecting error Web pages, and a multi-class Web subject-based categorization, both problems using convolutional neural networks.
\end{abstract}


\section{Introduction}
\label{sec:introduction}
Imagining today's world without the Internet is difficult since human activities such as commerce, education, entertainment, social interaction, and many more have their digital version. It is the tool that has allowed much of these activities to take place despite the paralysis caused by the recent pandemic. Internet is commonly associated with the \textit{Web}, both terms are intimately related but are very different concepts. The first one refers to the big network of networks, i.e., the infrastructure, whereas the second one refers to the content, formed by \textit{Web sites}, which are a collection of \textit{Web pages} on a specific topic and linked to each other \cite{DeBoer}.

Since its invention in the 1990s, the \textit{World Wide Web}, or just \textit{Web}, has revolutionized access to large amounts of data and information for individuals and organizations in all areas. Factors such as ease of use (user-friendly interface), popularity, and increasing connectivity have made the Web a field of interest for the business sector and scientific research. Among the main domains of development are the following:

\begin{itemize}
    \item Web design and redesign: identification of metrics and guidelines to support the work of beginners and experts.
    \item Analysis of aesthetics and quality of Web pages.
     \item Optimization and improvement of the indexing of Web pages by search engines as Google and Bing \cite{Du}.
    \item Categorization Web: classifiers and recommenders systems, Web directories and crawlers \cite{Lassri}.
    \item Security: detection of illegitimate Web sites (phishing).
    \item Accessibility without limitations, regardless of knowledge, skills and technology.
    \item Programming: automatic code generation.
    \item Machine Learning and Artificial Intelligence projects.
    \item Challenges of performance of algorithms and competitions of the best Web sites.
\end{itemize}

All of the above topics demand an essential resource: a large amount of data for analysis and testing tasks. Currently there is no large Web-related dataset that includes both visual representation and attributes of Web pages. Therefore, we make the following contributions:

\begin{itemize}
    \item A free and available dataset of 49,438 Web pages from all countries worldwide and classified in the following topics: arts and entertainment, business and economy, education, government, news and media, and science and environment. The dataset combines different types of data: images, text, and numbers, which represent the visual aspect of the full Web page (\textit{webshot}), and qualitative and quantitative attributes, respectively.
    \item A workflow supported by programming languages and computer tools to automatize most of the process of collecting, organizing, and debugging links, webshots, and parameter extraction. This methodology can be adapted to other problems where the acquisition of a lot of data is needed.
    \item A deep convolutional neural network to detect error Web pages using only their screenshots. This allowed us to debug our image dataset after the HTML code is no longer available. It can also be useful in other situations, e.g. for the job of crawlers, indexers and search engines by skipping analysis of HTML content, or webmasters could be notified of problems and hacking through periodic screenshots.
\end{itemize}

The rest of this paper is organized as follows: a review of similar works; a description of our dataset and the methodology used for its creation, debugging, and publication; a statistical analysis of the Web page parameters; and a case study on Web categorization to demonstrate the practical use of the dataset presented here.

\section{Related work}
\label{sec:state-of-the-art}
The availability of a large amount of data is the current need for research and development in areas related to the Web. We have generated an extensive dataset of Web pages, including features of various types: text, numbers, and images. First, we reviewed the literature and analyzed the existing datasets, whose main properties are in Table \ref{table:related_work_dataset}.

\begin{table*}[!ht]
\caption{State-of-the-art datasets.} \label{table:related_work_dataset}
\centering
\begin{tabular}{llllll}

\textbf{Owner \& year}                                                                          & \textbf{Size}                                                                     & \textbf{Topic}                                                                                            & \textbf{Data type}                                                                                                   & \textbf{Purpose}                                                                                             \\
\hline

\begin{tabular}[c]{@{}l@{}}De Boer\\ {\em et al.}, 2011\end{tabular}
                                                                                           & \begin{tabular}[c]{@{}l@{}}Small:\\ 60 screenshots\end{tabular}                     & \begin{tabular}[c]{@{}l@{}}News, hotels,\\ conferences, and\\ celebrities\end{tabular}              & \begin{tabular}[c]{@{}l@{}}Images\\ database\end{tabular}                                                                                                          & \begin{tabular}[c]{@{}l@{}}Aesthetics and thematic\\ classification with\\ Machine Learning\end{tabular}
\\
\begin{tabular}[c]{@{}l@{}}Reinecke\\ {\em et al.}, 2014\end{tabular}     & \begin{tabular}[c]{@{}l@{}}Small:\\ 430 screenshots\end{tabular}                   & Generic                                                                                                      & \begin{tabular}[c]{@{}l@{}}Images\\ database\end{tabular}                                                                                                          & \begin{tabular}[c]{@{}l@{}}Aesthetics\\ classification\end{tabular}                                               \\
\begin{tabular}[c]{@{}l@{}}López\\ {\em et al.}, 2017\end{tabular}
                                & \begin{tabular}[c]{@{}l@{}}Small:\\ 280 Web pages\end{tabular}                   & \begin{tabular}[c]{@{}l@{}}Food, animals,\\fashion, nature,\\ home and vehicles\end{tabular}              & \begin{tabular}[c]{@{}l@{}}URL and\\images extracted\\ from HTML\end{tabular} & \begin{tabular}[c]{@{}l@{}}Thematic \\ classification with\\ Machine Learning\end{tabular}                        \\

\begin{tabular}[c]{@{}l@{}}López\\ {\em et al.}, 2019\end{tabular}
      & \begin{tabular}[c]{@{}l@{}}Small:\\ 365 Web pages\end{tabular}                   & \begin{tabular}[c]{@{}l@{}}Food, vehicles,\\ animals, fashion,\\ home design and\\ landscape\end{tabular} & \begin{tabular}[c]{@{}l@{}}URL and\\images extracted\\ from HTML\end{tabular}             & \begin{tabular}[c]{@{}l@{}}Thematic \\ classification with\\ Machine Learning\end{tabular}                        \\
\begin{tabular}[c]{@{}l@{}}CIRCL, 2019\end{tabular} & \begin{tabular}[c]{@{}l@{}}Small:\\ 460 screenshots\end{tabular}                   & Phishing                                                                                                     & \begin{tabular}[c]{@{}l@{}}Images\\ database\end{tabular}                                                                                                          & \begin{tabular}[c]{@{}l@{}}Analysis of\\ security events\end{tabular}                                   \\

\hline
\begin{tabular}[c]{@{}l@{}}ImageNet, 2009\end{tabular}
                                                                                           & \begin{tabular}[c]{@{}l@{}}Large:\\ 1840 screenshots\end{tabular}                     & \begin{tabular}[c]{@{}l@{}}Generic\end{tabular}              & \begin{tabular}[c]{@{}l@{}}Images\\ database\end{tabular}                                                                                                          & \begin{tabular}[c]{@{}l@{}}Resource for\\image and vision\\ research field\end{tabular}                        \\
                                                                                           
\begin{tabular}[c]{@{}l@{}}Nordhoff\\ {\em et al.}, 2018\end{tabular}
      & \begin{tabular}[c]{@{}l@{}}Large:\\ 80901 screenshots\end{tabular}                                                                             & Generic                                                                                                      & \begin{tabular}[c]{@{}l@{}}URL,\\metrics\\and images\end{tabular}                                        & \begin{tabular}[c]{@{}l@{}}Aesthetics and\\ Web design\end{tabular}                        \\
\begin{tabular}[c]{@{}l@{}}CIRCL, 2019\end{tabular} & \begin{tabular}[c]{@{}l@{}}Large:\\ 37500 screenshots\end{tabular}                & \begin{tabular}[c]{@{}l@{}}Onion Website\\(Hidden Web,\\no indexed)\end{tabular}                                                                                                & \begin{tabular}[c]{@{}l@{}}Images\\ database\end{tabular}                                                                                                          & \begin{tabular}[c]{@{}l@{}}Analysis of\\security events\end{tabular} \\

\begin{tabular}[c]{@{}l@{}}University of\\Alicante, 2019\end{tabular}     & \begin{tabular}[c]{@{}l@{}}Large:\\ 8950 labeled\\ screenshots\end{tabular} & \begin{tabular}[c]{@{}l@{}}Good and\\ bad design\end{tabular}                                           & \begin{tabular}[c]{@{}l@{}}Labeled\\images\\dataset\end{tabular}                                                                                                          & \begin{tabular}[c]{@{}l@{}}Aesthetics Web\\ categorization\end{tabular}   \\

\hline
\end{tabular}

\end{table*}

For ease of comparison, we have divided the state-of-the-art datasets into two groups according to size: small (less than 1000 instances) and large. Other relevant properties are the topic, data types, its owner, purpose, and availability.

DeBoer \emph{et al.} \cite{DeBoer} use a tiny dataset, only visual and collected for categorization within four classes (news, hotels, conferences, and celebrities). These categories are quite different from each other, so the categorization problem is of less complexity. There is no link to download the screenshots.

Lopez \emph{et al.} \cite{Lopez} and \cite{Lopez2} have datasets with more Web pages, including their respective images and links (URLs). However, these images are not screenshots but elements of the Web page. The URL is used to download the images from HTML code and analyze them for categorization. Although there are more categories than in the previous work, they are still very different topics. In both cases, no download link.

Reinecke \emph{et al.} \cite{Reinecke} is the most relevant dataset among the smaller ones. It could be a useful resource for small-scale research and development works. It covers several countries in the world, various topics and is available for download. However, it is strictly visual and insufficient for current needs, and its purpose is more oriented to aesthetics analysis and classification.

The Computer Incident Response Center Luxembourg (CIRCL) is a government initiative created to respond to computer security threats and incidents. CIRCL \cite{CIRCL1} offers a dataset of more than 400 screenshots of verified or potential phishing Web sites. Also, an extensive dataset with more than 37000 images is available \cite{CIRCL2}, corresponding to screenshots of Web sites belonging to the Dark-Web, the problematic facet of the Web associated with cybercrime, hate, and extremism \cite{Fu}. Both datasets can be easily downloaded; however, because the images represent fakes or hidden Web pages would have limited applications.

ImageNet \cite{Imagenet}, the most popular of the image databases, includes millions of images organized according to the WordNet hierarchy\footnote{A large lexical database of English. Nouns, verbs, adjectives and adverbs are grouped into sets of cognitive synonyms (synsets).} (\href{https://wordnet.princeton.edu}{https://wordnet.princeton.edu}). The Web sites section has 1840 screenshots from different countries and languages without categorization. Some screenshots appear cropped, and download requires registration and authorization.

The dataset created by the University of Alicante \cite{UA} collects 8950 screenshots of Web pages for analysis and evaluation of the quality of Web design. Half of the images come from the Awwwards site (\url{https://www.awwwards.com}), so they have been labeled with "good design", whereas the other half extracted from yellow pages, labeled with "bad design". This dataset serves the academic work of the institution.

Nordhoff \emph{et al.} \cite{Nordhoff} stands out because it has achieved a larger number of Web pages. However, they come from only 44 countries, the parameters are purely aesthetics, and the image download is not direct. In contrast, our dataset takes into account all countries worldwide, includes parameters related to Web page structure, is general-purpose, and available for download.

According to the above, it is not possible to take advantage of existing data sources, so our purpose is to create from scratch an extensive and available dataset. It incorporates the visual representation of the Web page through a webshot, complemented with qualitative and quantitative parameters extracted from the underlying HTML source code, so that a Web page is better characterized. Because manual data collection is a complex task and requires too much time and human effort, we have automatized most of the process by writing several programs in Python and R.

\section{Description of the dataset}
\label{sec:dataset}
The dataset has been designed to combine visual, textual and numerical elements, which are presented in Table~\ref{table:variables} and detailed below.

\begin{table}[H]
\caption{Structure of the dataset.}
\label{table:variables}
\centering
\begin{tabular}{lcl}
\textbf{Element} & \textbf{Type}                  & \textbf{Description}                                                                           \\
\hline
Webshot        & Visual                         & Web page entire screenshot in JPG format                                          \\
\hline
Name        & Text                         & Identification given to the webshot                                          \\
\hline
URL               & Text                        & Link to locate and display a Web page                            \\
\hline
Country              & \multirow{3}{*}{Qualitative}   & National origin of Web page                    \\
Continent        &                                & Region grouping countries                                                       \\
Category         &                                & Main thematic of Web page                                             \\
\hline
Time              & \multirow{11}{*}{Quantitative} & Web page's source code download time                                 \\
Bytes             &                                & Size in bytes of Web page's source code                                                       \\
Images            &                                & Number of images from Web page                                                \\
Script\_files     &                                & Number of executable files of Web page                \\
CSS\_files        &                                & Number of files to layout a Web page \\
Tables            &                                & Number of \textit{table} tags in the source code                                \\
iframes           &                                & Number of \textit{iframe} tags in the source code                                       \\
Style\_tags       &                                & Number of \textit{style} tags in the source code                                        \\
Img\_bytes        &                                & Webshot size in bytes                                              \\
Img\_width        &                                & Webshot width in pixels                                                \\
Img\_height       &                                & Webshot height in pixels         \\
\hline
\end{tabular}
\end{table}

A \textbf{webshot} is a digital image of the entire Web page; unlike a \textit{screenshot}, which may appear cropped because its dimensions exceed the viewing device, forcing the user to scroll. The \textbf{name} given to the webshot is a key element that follows a convention to identify the Web page's source, category, and country. It is also the link between the image and the qualitative and quantitative parameters. A \textbf{URL} (Uniform Resource Locator) is the Web page's address together the recovery mechanism (\textit{http} / \textit{https}). It is placed in the address bar of the \textit{browsers}, which are the programs to display the content to the user. We have collected URLs worldwide to cover different cultural characteristics and preferences, so the dataset includes attributes related to the geographic location such as \textbf{country} and \textbf{continent}. The Web pages collected belong to the following \textbf{categories}: Arts and Entertainment, Business and Economy, Education, Government, News and Media, and Science and Environment. We have considered these categories since they are part of the Web directory used here and explained in section \ref{subsubsec:browsing}.

We have added the following quantitative parameters, which give us an overview of the structure and quality of a Web page:

\begin{itemize}
    \item \textbf{Download time}: users want to wait as little as possible to view a Web page \cite{King}, which means reducing the source code download time.
    
    \item \textbf{Size}: the larger the size in bytes, the slower the download and display of the Web page.
    
    \item \textbf{Images}: the images will increase the download time. A Web page is not more attractive because it has more images, a balance between all types of information is recommendable.

    \item \textbf{Scripts}: they are external files to get more complex functionality to the Web page. It is convenient to reduce their quantity because they increase the network traffic and download time.

    \item \textbf{CSS files}: style files cause an extra load and delay the display of the Web page, ideally there should be one.

    \item \textbf{Tables}: are often used to structure the Web page's content; however, it is discouraged due to appropriate elements such as "div" tags.

    \item \textbf{iFrames}: insert a Web page inside another one, which is not a good practice currently.

    \item \textbf{Style tags}: are not recommended since there are CSS files.

\end{itemize}

\begin{figure}[H]
\centering
    \includegraphics[width=0.99\textwidth]{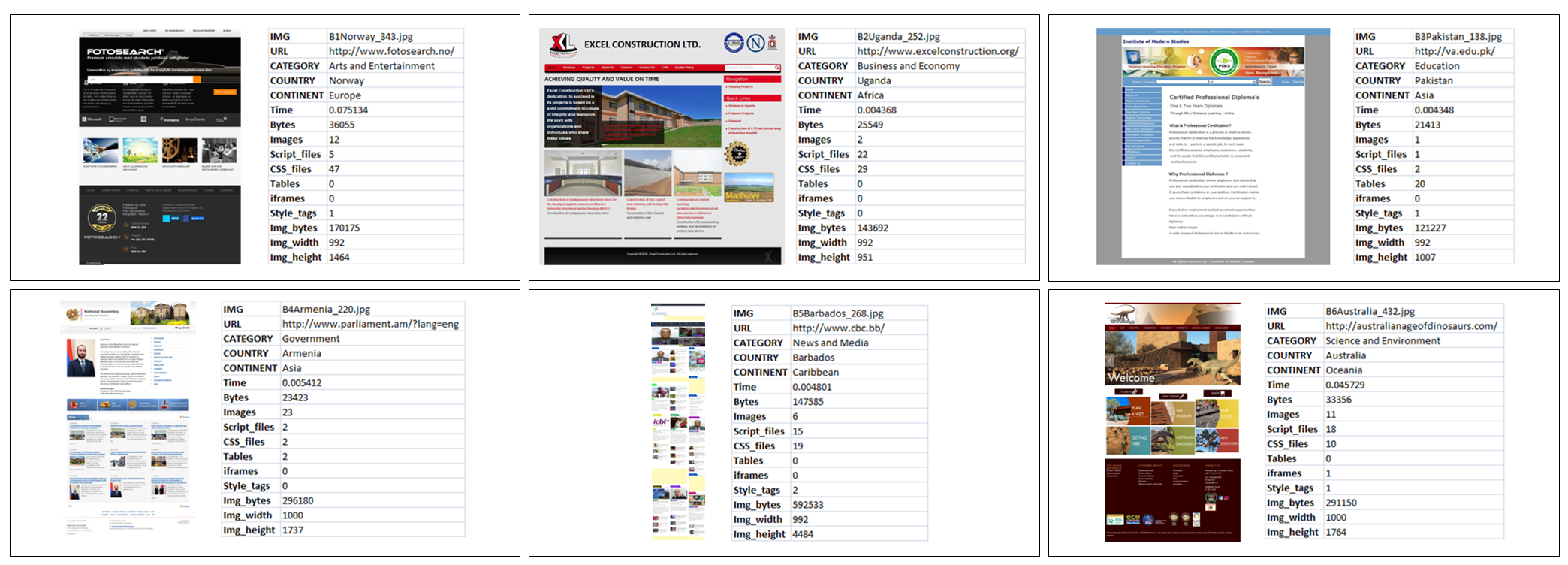}
    \caption{\label{fig:sample} A small sample of the dataset, including one example of each category.}
\end{figure}

Finally, we have the \textbf{image size} in bytes, as well as the dimensions (\textbf{width} and \textbf{height}) in pixels of each webshot. Fig.~\ref{fig:sample} presents a small sample of the dataset showing a case of each category.

\section{Methodology}
\label{sec:methodology}

\begin{figure}[!ht]
\centering
    \includegraphics[width=0.65\textwidth]{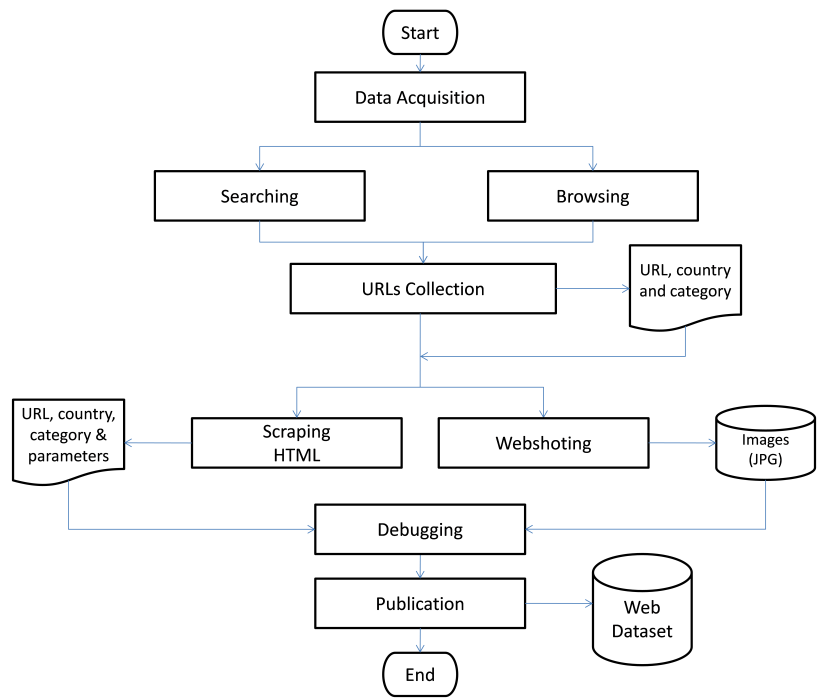}
    \caption{\label{fig:methodology} Methodology created to obtain the Web pages dataset.}
\end{figure}

The workflow developed to produce the dataset is represented in Fig.~\ref{fig:methodology}. Following, we describe each stage included in the methodology, supported by various techniques, procedures, and computer tools.

\subsection{Data Acquisition}
\label{subsec: acquire}
The first step is to collect URLs worldwide related to the given categories. Then, each of the URLs will allow us to download the HTML code, extract quantitative and qualitative parameters by scraping, and take a screenshot of the entire Web page (webshot).

In order to obtain a larger number of URLs, we have used two ways of finding information on the Web: \textit{"Searching"} and \textit{"Browsing"}. Searching requires the user to translate a need for information into queries, whereas Browsing is a basic and natural human activity, occurring in an information environment where information objects are visible and organized \cite{Ruthven}. 
Below, we describe in detail how both techniques allowed us to collect URLs, and based on them, extract parameters and capture webshots, all through various Python and R scripts.

\subsubsection{URL collection by Searching}
\label{subsubsec:searching}

The Google search engine asks for words or phrases related to the topic of interest. To avoid the repetitive task of typing the query into the Google search page and manually retrieving the response URLs, we have automatized the process through a Python script\footnote{\url{https://osf.io/k6yrx}}, where:

\begin{enumerate}
    \item The country name and its Internet code are extracted iteratively from a plain text file\footnote{\url{https://osf.io/yrmx8}}.
    \item The search query has the following structure:
    
    \scriptsize{'site:' + \textit{countrycode} + ' business OR economy OR marketing OR computers OR internet OR construction OR financial OR industry OR shopping OR restaurant' + ' ext:html'}
    \normalsize
    
    \textit{OR}, \textit{site} and \textit{ext} are operators or reserved words that can be used in query phrases within the Google search engine. The "OR" operator concatenates several search words related to the category. The "site" operator specifies the geographic top-level Internet domain assigned for each country, e.g. ".es" for Spain. The "ext:html" operator allows us to obtain results exclusively with this extension.
    \item The request returns the Google results page with the first 100 links. We have defined it like this to achieve an approximately uniform distribution of Web pages according to country and category.
    \item Web page links are extracted by automatically scanning the source code of the results page (\textit{scraping}), generating a text file that contains the URLs and their attributes: country, continent and category.
\end{enumerate}

For the rest of the categories, the query phrases are:

\vspace{0.25cm}

\noindent
\scriptsize{'site:' + \textit{countrycode} + ' arts OR entertainment OR dance OR museums OR theatre OR literature OR artists OR galleries' + ' ext:html'}

\vspace{0.20cm}
\noindent
\scriptsize{'site:' + \textit{countrycode} + ' education OR academy OR university OR college OR school' + ' ext:html'}

\vspace{0.20cm}
\noindent
\scriptsize{'site:' + \textit{countrycode} + ' government OR military OR presidency ' + ' ext:html'}

\vspace{0.20cm}
\noindent
\scriptsize{'site:' + \textit{countrycode} + ' news OR media OR magazine OR radio OR television OR newspaper ' + ' ext:html'}

\vspace{0.20cm}
\noindent
\scriptsize{'site:' + \textit{countrycode} + ' science OR environment OR archaeology' + ' ext:html'}
\normalsize

\subsubsection{URL collection by Browsing}
\label{subsubsec:browsing}
This technique uses a \textit{Web Directory}, a specialized Web site consisting of a catalog of links to other Web sites. Build, maintain, and organization by categories and subcategories is done by human experts, unlike search engines that do it automatically. To include a URL, specialists perform a review, analysis, and evaluation process to verify the requirements determined by the Web Directory.

\begin{figure}[H]
    \centering
    \fbox{\includegraphics[width=0.5\textwidth]{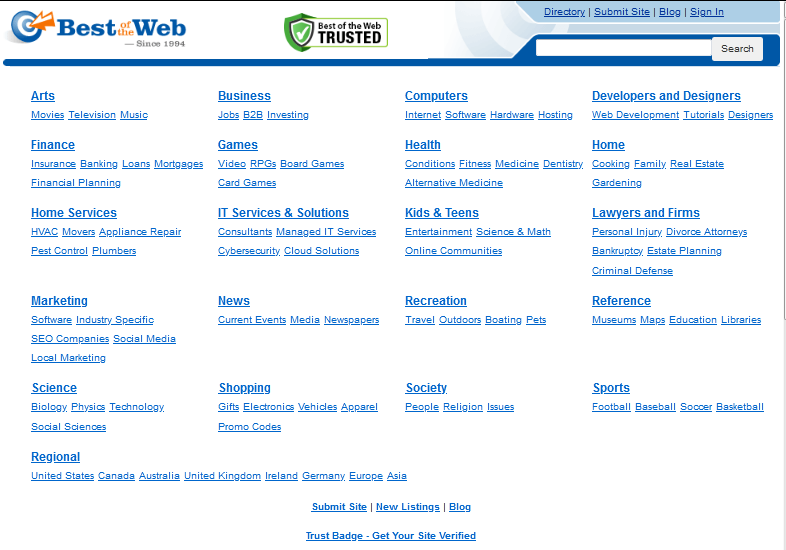}}
    \caption{\label{fig:botw} BOTW Web Directory (\url{https://botw.org/})}
\end{figure}

Today a few Web Directories have survived the popularity of search engines like Google. We can highlight \textit{Best of the Web (BOTW)} (Fig.~\ref{fig:botw}), one of the most recognized by its quality, global reach, a wide range of categories and subcategories, level of traffic (visits per month), reliability, the number of links, and demanding requirements.

Instead of a query or search phrase, it is necessary to know the hierarchical structure of the directories and subdirectories until the URL of interest. We have taken advantage of the organization by countries and categories defined by BOTW Web Directory, e.g. for Greece:

\vspace{0.2cm}
\tiny
\begin{verbatim}
https://botw.org/top/Regional/Europe/Greece/Arts-and-Entertainment/
https://botw.org/top/Regional/Europe/Greece/Business-and-Economy/
https://botw.org/top/Regional/Europe/Greece/Education/
https://botw.org/top/Regional/Europe/Greece/Government/
https://botw.org/top/Regional/Europe/Greece/News-and-Media/
https://botw.org/top/Regional/Europe/Greece/Science-and-Environment/
\end{verbatim}
\normalsize
\vspace{0.2cm}

We collect the URLs published within each category through a Python script\footnote{\url{https://osf.io/73sc2}} that:

\begin{enumerate}
    \item Reads iteratively the name of each country from a flat text file\footnote{\url{https://osf.io/ve986}}.
    \item Sets the path corresponding to the category, which will always have the same structure, only the country's name changes:\\
    \small
    'https://botw.org/top/Regional/' + \textit{countryname} + '/Science\_and\_Environment/'
    \normalsize
    \item A connection to the formed Web address is realized, which obtains the source code of the results page to extract the URLs of each category.
    \item The links are stored in a text file\footnote{\url{https://osf.io/hjwgm}} that can be opened in a spreadsheet where a filter is applied to select only those links belonging to a country in particular.
\end{enumerate}

\subsubsection{Parameters collection by Scraping}
\label{subsubsec:scraping}
After collecting and storing the URLs from the two sources described above, we implemented a Python script\footnote{\url{https://osf.io/de78f}} to: a) sequentially read the URL links stored in the text file\footnote{\url{https://osf.io/gk3p2}}; b) make a connection via browser to each of these links; and c) download and analyze the source code of the Web page, obtaining the parameters specified in the dataset: download time in seconds, total size in bytes, number of images, script files, CSS files, tables, iFrames tags, and style tags. This script includes the library for Web scraping named \textit{Beautiful Soup}, which allows us to define and extract the mentioned parameters from the HTML code of each Web page.

\subsubsection{Webshots collection}
\label{subsubsec:webshots}
We take advantage of \textit{Webshot} and \textit{PhantomJS} packages to create a R script\footnote{\url{https://osf.io/6pmyb}} that: reads each URL from text files generated in sections \ref{subsubsec:searching} and \ref{subsubsec:browsing}, takes a snapshot of the entire Web page, and saves it as JPG image. 

The name assigned to the image is an unique identifier that allows us to know the source, category, and country to which the Web page belongs, e.g. \textit{B2Netherlands\_791.jpg} indicates a screenshot of a Web page obtained through the technique of Browsing, belonging to category number two (Business and Economy) and that it comes from the Netherlands. The number after the underscore only establishes a sequential order. Note that such identifier is the key to attach the webshots to their respective qualitative and quantitative parameters. In this way, it was possible to link two different storage media, i.e. a parameters datasheet and an images folder.

Additionally, this script includes functions to obtain both webshot size in bytes and webshot dimensions (width and height) in pixels, so we get a better characterization of the Web page. All images are available for viewing and downloading\footnote{\url{https://osf.io/7ghd2/?view_only=0bf99589809e4e88b0aa0602c8060b46}}.

\subsection{Dataset debugging}
\label{subsec:debugging}
During the automatic data collection, some events did not allow downloading the HTML code or capturing the webshot, caused by: request for manual acceptance of \textit{cookies} and \textit{SSL} certificates, error messages as \textit{HTTP 403 Forbidden}, \textit{HTTP 404 Not Found}, \textit{HTTP 406 Not Acceptable}, \textit{HTTP 909 Denied permission}, and exceed timeout.

We used exception handling inside the scripts to avoid interruptions during the execution of the programs. When an error occurred, the fields associated with the parameters or webshot were assigned the value "-1". Thus, the programs could continue their execution, and the inexistence of webshots or parameters was solved.

\begin{figure}[!ht]
\centering
    \includegraphics[width=0.75\textwidth]{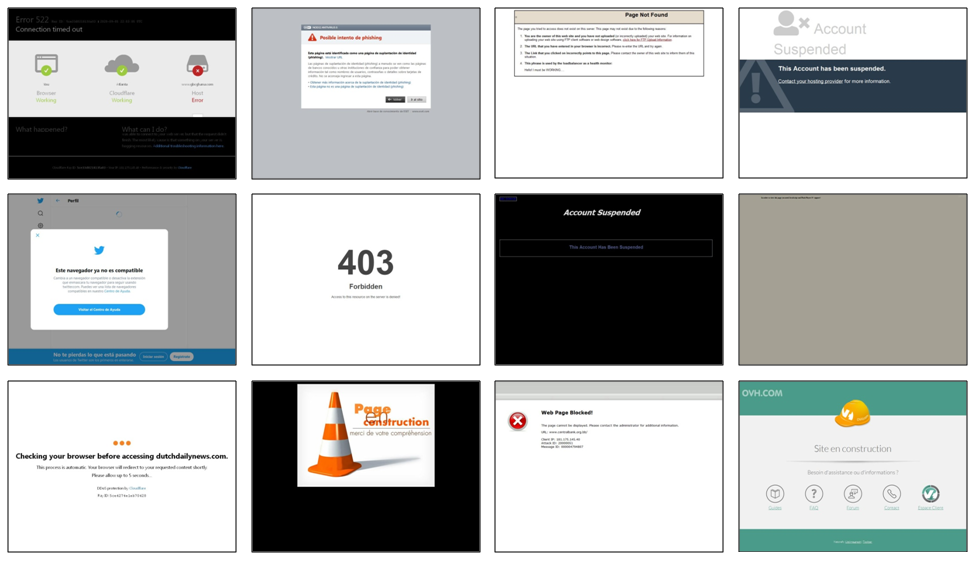}
    \caption{\label{fig:errorpages} A sample of the error Web pages.}
\end{figure}

For the final dataset, we have considered only URLs that have their respective webshot, as this is the most relevant element of our work. However, after a brief visual review of the webshots in the dataset, several error Web pages were detected, e.g. Web sites under construction, maintenance, domain offer, suspended account, page not found, browser incompatibility, virus or phishing risks. Some of them shown in Fig.~\ref{fig:errorpages}.

The webshots above are not useful for the dataset, so we decided to select and remove these images that appeared in the automatic collection. Although the size of the final dataset will be smaller, we will obtain a cleaner dataset. Given that the URLs connections corresponding to these webshots did not return HTTP 403 or HTTP 404 error messages, nor did the HTML code contain phrases like "suspended account" or "page under construction", text analysis was not possible. We need implement an image analyzer to avoid manual and visual verification of thousands of webshots, which consumes too much time and effort. We used a \textit{Convolutional Neural Network} (CNN), the state-of-the-art tool in computer vision, to detect error Web pages, and then separate them into a dedicated folder, all automatically.

\begin{figure}[!ht]
\centering
    \includegraphics[width=0.5\textwidth]{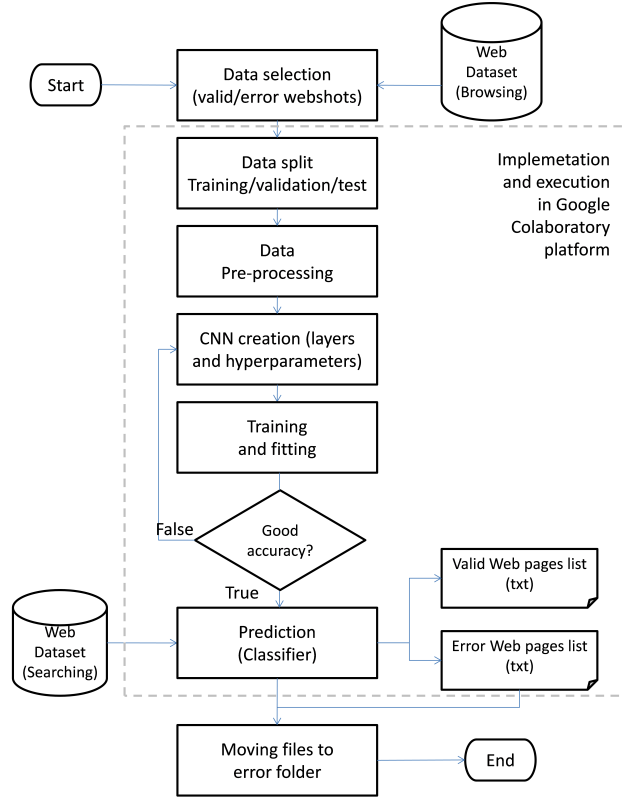}
    \caption{\label{fig:dfderror} Methodology for detecting error Web pages.}
\end{figure}

Here we present an automatic detection of error Web pages based exclusively on their webshots. It consists of determining if a Web page belongs to a "VALID" category or, by the contrary, to an "ERROR" category, i.e. a binary classification problem. To tackle it, we followed the methodology shown in Fig.~\ref{fig:dfderror}.

\subsubsection{Data selection}
\label{subsubsec:dataselection}
The main resource for automatic learning is data. In our case, images that will be the input for a training process that tries, iteratively, to obtain a known output ("valid" or "error"), and if an acceptable accuracy is reached, to make predictions.

The training process requires images associated with their respective category: valid or error. Since our dataset consists of two groups of images (Browsing and Searching), we selected the webshots of the smallest subset (Browsing) to perform an exhaustive visual inspection and classify the images manually, obtaining the results shown in Table \ref{table:BrowsingTable}. Once the neural network model is adjusted, it will classify each webshot in the largest subset (Searching) as valid or error.

\begin{table}[!ht]
\centering
\caption{Dataset for binary classification (Browsing webshots). \label{table:BrowsingTable}}

\scalebox{0.9}{

\begin{tabular}{c|c|c|c|}
\cline{2-4}
                                                & \multicolumn{3}{c|}{\textbf{Browsing}}          \\ \hline
\multicolumn{1}{|c|}{\textbf{Category}}         & \textbf{Webshots} & \textbf{Valid} & \textbf{Error} \\ \hline
\multicolumn{1}{|c|}{Arts \& Entertainment}  & 447           & 397            & 50             \\ \hline
\multicolumn{1}{|c|}{Business \& Economy}    & 1058          & 892            & 166            \\ \hline
\multicolumn{1}{|c|}{Education}                 & 419           & 368            & 51             \\ \hline
\multicolumn{1}{|c|}{Government}                & 730           & 669            & 61             \\ \hline
\multicolumn{1}{|c|}{News \& Media}            & 458           & 394            & 64             \\ \hline
\multicolumn{1}{|c|}{Science \& Environment} & 497           & 462            & 35             \\ \hline
\multicolumn{1}{|c|}{\textbf{Total}}            & \textbf{3609} & \textbf{3182}  & \textbf{427}   \\ \hline
\end{tabular}
}
\end{table}

\begin{figure}[H]
\centering
    \includegraphics[width=0.20\textwidth]{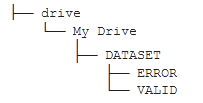}
    \caption{\label{fig:dir_structure_error} Directory structure of the dataset.}
\end{figure}

The dataset for training an error Web pages detection model has a total of 3609 images, 427 error webshots and 3182 valid webshots, which have been uploaded to \textit{Google Drive} in separated folders, called "VALID" and "ERROR" (Fig.~\ref{fig:dir_structure_error}).

We take advantage of \textit{Google Colaboratory}, a free platform that offers powerful hardware and requires no installation or setup, supports Python through an online notebook, and includes the packages and libraries to facilitate automatic learning such as \textit{Tensorflow}, \textit{Keras}, \textit{Sklearn}, and others. Next, a description of the developed code\footnote{\url{https://colab.research.google.com/drive/1mzmm0C-WDNopOlEtRm7VVA6_8eHuYoVH?usp=sharing}}.

The initial step is the connection to the data source where the images of our dataset have been stored within folders and subfolders, named as the categories (Fig.~\ref{fig:dir_structure_error}). Such denomination facilitates the labeling of the images with their corresponding category. A pair of instructions are needed to access \textit{Google Drive}:

\scriptsize
\begin{verbatim}
from google.colab import drive
drive.mount('/content/drive')
\end{verbatim}
\normalsize

\subsubsection{Split data}
\label{subsubsec:splitdata}
One of the tasks that characterizes automatic learning is the division of the data. Because we have only 3609 images, we consider two subsets: training and validation (Table \ref{table:DatasetSplit}). The training subset contains the largest amount of images (80\%) and is used for learning and fitting the model's parameters, whereas the validation subset (20\%) is used to evaluate the capacity of the model, which will serve to adjust its \textit{hyperparameters}.

\begin{table}[!ht]
\centering
\caption{Dataset split for training and validation. \label{table:DatasetSplit}}
\scalebox{0.9}{
\begin{tabular}{|c|c|c|c|c|}
\hline
\textbf{Subset} & \textbf{Webshots} & \textbf{Valid} & \textbf{Error} & \textbf{Percentage} \\ \hline
Training        & 2886              & 2545           & 341            & 80\%                \\ \hline
Validation      & 723               & 637            & 86             & 20\%                \\ \hline
\textbf{Total}  & 3609              & 3182           & 427            & 100\%               \\ \hline
\end{tabular}
}
\end{table}

Although the most convenient is a balanced dataset, that is, an equal number of error cases and valid cases, we used all the images in order to obtain a better generalization. To automatically divide into training and validation folders, is useful to install and import the \textit{split-folders}\footnote{\href{https://pypi.org/project/split-folders/}{https://pypi.org/project/split-folders/}} package. It is necessary to specify the images directory, output directory, and the proportion to split (80\% and 20\%, respectively).

\vspace{0.20cm}

\scriptsize
\begin{verbatim}
splitfolders.ratio('/content/drive/My Drive/DATASET',
    output='/content/drive/My Drive/SPLIT', seed=1337,
    ratio=(.8,.2), group_prefix=None) 
\end{verbatim}
\normalsize

\vspace{0.20cm}

The result is a new directory structure. Within the "SPLIT" folder, the "train" and "val" folders are created, and within each of these, the "ERROR" and "VALID" folders.

\subsubsection{Data pre-processing}
\label{subsubsec:preprocessing}
The images need to be prepared before modeling. First, we normalize the pixel values (integers between 0 and 255) to a scale between 0 and 1, using an image processing utility. The \textit{ImageDataGenerator} class from Keras framework divides all pixels values by the maximum pixel value (255).

\vspace{0.20cm}

\scriptsize
\begin{verbatim}
train_datagen = ImageDataGenerator(rescale=1./255)
\end{verbatim}
\normalsize

\vspace{0.20cm}

Second, the images have different dimensions (width and height), so they are all resized to 256x256 pixels by setting the \textit{target\_size} parameter of the \textit{flow\_from\_directory} method. This operation is performed in groups of 32 images (\textit{batch\_size}) that are labeled for binary classification (\textit{class\_mode}) according to the folder where they are stored (valid and error) within of training directory.

\vspace{0.20cm}

\scriptsize
 \begin{verbatim}
train_generator = train_datagen.flow_from_directory(
    train_data_dir,
    target_size=(256,256),
    batch_size=32,
    shuffle=False,
    class_mode='binary')
\end{verbatim}
 \normalsize

\vspace{0.20cm}
 
 In this way, the small values of both pixels and dimensions help speed up the training process. The above code applies to the validation data, only the directory changes.

\subsubsection{Create model}
\label{subsubsec:model}
The model's architecture is based on the convolutional neural network proposed by Liu \emph{et al.} \cite{Liu} to detect malicious Web sites. Since this is a similar problem, we just applied minor adaptations. The model's structure (Fig.~\ref{fig:model}) is composed of two parts:

\begin{figure*}[!ht]
\centering
    \includegraphics[width=0.99\textwidth]{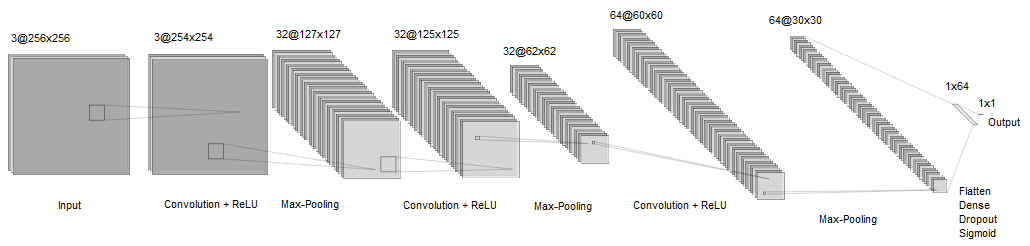}
    \caption{\label{fig:model} CNN architecture.}
\end{figure*}

\begin{itemize}
    \item Convolutional basis: for automatic extraction of features. The input image is resized to 256x256 pixels and separated into 3 RGB color channels (Red, Green, Blue). Then, it is processed by 3 convolutional layers with their respective activation functions (ReLU, Rectified Linear Unit) and max-pooling layers. The first two convolutions use 32 filters (kernels) while the third convolution has 64, with a size of 3x3, in contrast to the pool size is 2x2.
    \item Binary classifier: features are received in a flattened form by a fully connected layer that applies dropout to avoid overfitting. The sigmoid function generates the prediction as a probability value between 0 and 1. If the value is greater than 0.5, the Web page is valid, otherwise, an error Web page. 
\end{itemize}

Keras provides functions to implement this model from scratch in a simple way, just adding in sequence the convolutional, activation, pooling, dropout, flatten, and dense layers, and specifying their respective parameters. For example, the first convolutional block has the following instructions:

\vspace{0.20cm}

\scriptsize
\begin{verbatim}
model = Sequential()
model.add(Conv2D(32, (3, 3), input_shape=(256, 256, 3)))
model.add(Activation('relu'))
model.add(MaxPooling2D(pool_size=(2, 2)))
\end{verbatim}
\normalsize

\vspace{0.20cm}

Meanwhile, the classifier part:

\vspace{0.20cm}

\scriptsize
\begin{verbatim}
model.add(Flatten())
model.add(Dense(64))
model.add(Activation('relu'))
model.add(Dropout(0.5))
model.add(Dense(1))
model.add(Activation('sigmoid'))
\end{verbatim}
\normalsize

\vspace{0.20cm}

\begin{figure}[H]
\centering
    \includegraphics[width=0.4\textwidth]{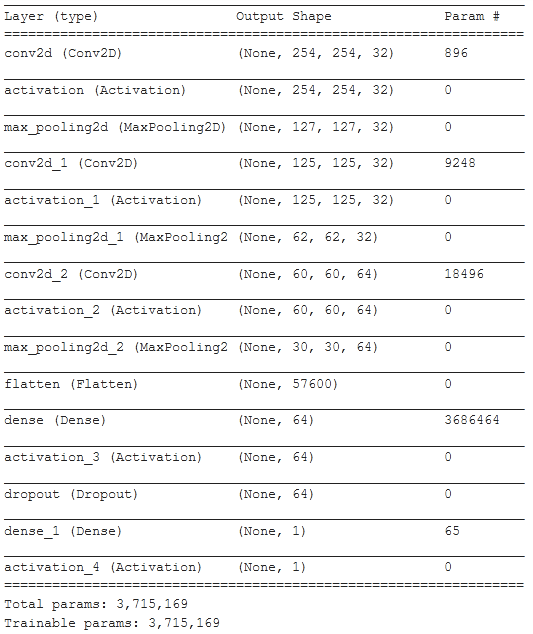}
    \caption{\label{fig:modelsummary} Model's layers summary.}
\end{figure}

Finally, the \textit{model.summary()} instruction allows us to verify the structure in detail (Fig.~\ref{fig:modelsummary}).

\subsubsection{Train and fit model}
\label{subsubsec:train}
Before starting the training, we must explicitly define the \textit{hyperparameters} required by the neural network for binary classification.

\vspace{0.20cm}

\scriptsize
\begin{verbatim}
model.compile(loss='binary_crossentropy',
    optimizer='rmsprop', metrics=['acc'])
\end{verbatim}
\normalsize

\vspace{0.20cm}

Through Keras, we can establish the loss function that will be minimized by the optimization algorithm and the classification accuracy as the metric that will be collected and reported by the model.

\vspace{0.20cm}

\scriptsize
\begin{verbatim}
history = model.fit(
    train_generator,
    steps_per_epoch=train_samples // batch_size,
    epochs=epochs,
    validation_data=val_generator,
    validation_steps=validation_samples // batch_size,
    callbacks=[checkpoint])
\end{verbatim}
\normalsize

\vspace{0.20cm}

\begin{figure}[H]
\centering
    \includegraphics[width=0.4\textwidth]{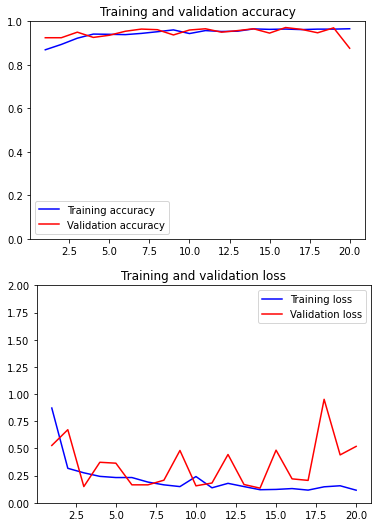}
    \caption{\label{fig:curves} Accuracy and loss in training and validation phases.}
\end{figure}

After a few hours of computation, 20 iterations (\textit{epochs}) of the training dataset (2886 images) have been executed; each iteration consists of 90 groups (\textit{steps\_per\_epoch}) of 32 images (\textit{batch\_size}). The accuracy achieved is \textbf{96.6\%} (iteration 20), while in the validation stage the accuracy is \textbf{97.16\%} (iteration 16). The evolution of the process is summarized in the graphs of the training and validation curves (Fig.~\ref{fig:curves}).

The training and validation phases have reached a high level of accuracy, both progressing to the same level, which is desirable. The model fits very well with the images provided, but how it will behave with new images (generalization). This concern is answered by analyzing the difference between training and validation losses. The last one, despite oscillating, does not separate so much from the other one until iteration 17, after which, they start to separate, with the possibility of overfitting. Therefore, the model is saved with the accuracy and parameters of iteration number 16. We can say that the model is capable of acceptably distinguishing error Web pages and valid Web pages and thereby move to the prediction phase.

\subsubsection{Predictions}
\label{subsubsec:predictions}
The images from the largest set (Searching) of our dataset become the input of the already trained and validated model. We used \textit{google.colab} library to select and upload from local drive the file (webshot) with click on “Choose Files” button.

\vspace{0.20cm}

\scriptsize
\begin{verbatim}
from google.colab import files
uploaded = files.upload()
\end{verbatim}
\normalsize

\vspace{0.20cm}

Once the file to be 100\% uploaded, it will be pre-processed using \textit{keras.preprocessing} and \textit{NumPy} libraries to transform the image into an array with a suitable shape and normalized pixel values for the model, which makes the prediction.

\vspace{0.20cm}

\scriptsize
 \begin{verbatim}
 img = image.load_img(path, target_size=(256, 256))
 x = image.img_to_array(img)/255.
 x = np.expand_dims(x, axis=0)
 images = np.vstack([x])
 classes = model.predict(images)
\end{verbatim}
\normalsize

\vspace{0.20cm}

\begin{figure}[H]
\centering
    \includegraphics[width=0.6\textwidth]{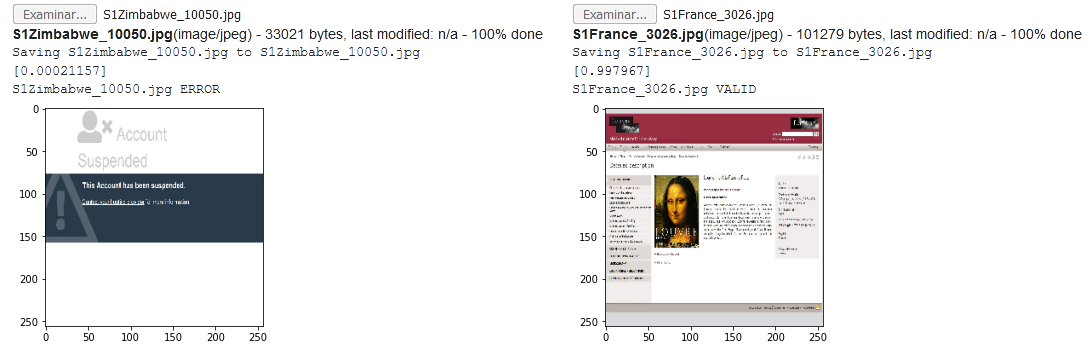}
    \caption{\label{fig:predictions} Prediction of error and valid Web page.}
\end{figure}

Fig.~\ref{fig:predictions} shows the result for two images selected one at a time. The resized webshot is displayed and the prediction is a probability value between 0 and 1, less than 0.5, so the category assigned is ERROR (left side), and a case of a valid Web page, with a probability value very close to 1 (right side).

\begin{figure}[!ht]
\centering
    \includegraphics[width=0.3\textwidth]{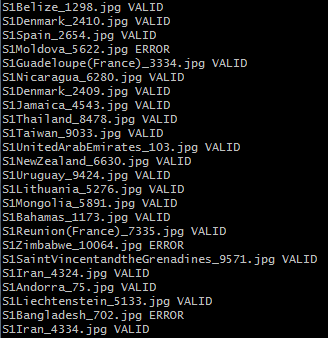}
    \caption{\label{fig:listsample} Prediction for a group of Web pages.}
\end{figure}

In addition to making predictions one by one, more important for our purpose is to generate predictions for groups of images. We just select a list of files using the choose button. As our dataset is organized by topic categories, we can select all images in one category, e.g. "Arts and Entertainment". An extract of the results is shown in Fig.~\ref{fig:listsample}.

This list of predictions is passed to a spreadsheet, then by means of a filter, the Web pages of the error category are selected and saved as a text format file (\textit{list.txt}). This file is the input to execute a command that moves all images from their original folder to the "ERROR" folder. This command-line runs in \textit{Windows} (PowerShell interface), although it is easily adaptable to different operating systems such as \textit{Linux}.

\vspace{0.20cm}
\scriptsize
\begin{verbatim}
cat list.txt | ForEach {mv $_ ERROR}
\end{verbatim}
\normalsize
\vspace{0.20cm}

As result of the prediction, 822 error Web pages y 7747 valid Web pages. Once the images were classified and separated, the visual verification was much faster, and we were able to manually establish the successes and fails of the classifier. Thus we identified 1214 real error Web pages and 7355 real valid Web pages. The same procedure is carried out for the remaining images in the other categories. The results are summarized in Table \ref{table:classificationresults}.

\begin{table}[H]
\centering
\caption{Results of the binary Web categorization.}
\label{table:classificationresults}
\scalebox{0.75}{
\begin{tabular}{c|c|c|c|c|c|c|}
\cline{2-7} & \multicolumn{6}{c|}{Searching} \\
\hline
\multicolumn{1}{|c|}{Category} & Webshots & 
\begin{tabular}[c]{@{}c@{}}Valid\\ (Prediction)\end{tabular} & \begin{tabular}[c]{@{}c@{}}Error\\ (Prediction)\end{tabular} &
\begin{tabular}[c]{@{}c@{}}Valid\\ (Real)\end{tabular} & \begin{tabular}[c]{@{}c@{}}Error\\ (Real)\end{tabular} & Accuracy \\
\hline
\multicolumn{1}{|c|}{\begin{tabular}[c]{@{}c@{}}Arts \&\\ Entertainment\end{tabular}}
  & 8569 & 7747 & 822 & 7355 & 1214 & 94.68\% \\
\hline
\multicolumn{1}{|c|}{\begin{tabular}[c]{@{}c@{}}Business \&\\ Economy\end{tabular}}
 & 8699 & 8004 & 695 & 7546 & 1153 & 93.79\% \\
\hline
\multicolumn{1}{|c|}{Education}              & 8742 & 8083 & 659 & 7524 & 1218 & 92.80\% \\[2ex]
\hline
\multicolumn{1}{|c|}{Government}             & 8088 & 7363 & 725 & 6685 & 1403 & 90.85\% \\[2ex]
\hline
\multicolumn{1}{|c|}{\begin{tabular}[c]{@{}c@{}}News \&\\ Media\end{tabular}}         & 11574 & 10597 & 977 & 9650 & 1924 & 90.80\% \\
\hline
\multicolumn{1}{|c|}{\begin{tabular}[c]{@{}c@{}}Science \&\\ Environment\end{tabular}}
 & 8893  & 8137  & 756 & 7496 & 1397 & 92.34\% \\
\hline
\multicolumn{1}{|c|}{Total}                  & 54565 & 49931 & 4634 & 46256 & 8309 & 92.47\% \\
\hline
\end{tabular}
}
\end{table}

We used the \textit{Confusion Matrix} to evaluate and know how accurate our model is. This table enables us to compare reality and prediction, and based on the successes and fails, calculates an accuracy value. E.g. for "Arts and Entertainment" category, the classifier predicted 822 error Web pages, however, it failed in 32 instances. There were 7747 predictions of valid Web page, but 424 were incorrect. These values are placed in Table \ref{table:cm1} and substituted into the formula for accuracy.

\begin{table}[!ht]
\centering
\caption{Confusion Matrix for the "Arts and Entertainment" category.}
\label{table:cm1}
\begin{tabular}{cc|c|c|}
\cline{3-4}
                                                      &       & \multicolumn{2}{c|}{\textbf{Prediction}}                    \\ \cline{3-4} 
                                                      &       & Error                        & Valid                        \\ \hline
\multicolumn{1}{|c|}{}                                & Error & 790 & 424                          \\ \cline{2-4} 
\multicolumn{1}{|c|}{\multirow{-2}{*}{\textbf{Real}}} & Valid & 32                          & 7323 \\ \hline
\end{tabular}
\end{table}

\begin{center}
$Accuracy = \frac{790+7323}{790+32+424+7393} = 0.9468 \times 100\% = 94.68\% $
\end{center}

\vspace{0.25cm}

The classifier reached an accuracy of 94.68\% for this category, which is good considering the small number of images that were part of the training process. Table \ref{table:confusionmatrixoverall} shows the respective confusion matrix for each category and a total matrix indicating an accuracy of \textbf{92.47\%} for the entire Searching dataset.

\begin{table}[!ht]
\centering
\caption{Confusion matrix for the rest of categories and overall result.}
\label{table:confusionmatrixoverall}
\begin{minipage}[b]{0.45\linewidth}
Business and Economy \\ \\
\centering
\scalebox{0.85}{
\begin{tabular}{cc|c|c|}
\cline{3-4} & & \multicolumn{2}{c|}{\textbf{Prediction}} \\
\cline{3-4} & & Error & Valid \\
\hline
\multicolumn{1}{|c|}{} & Error & 654 & 499 \\
\cline{2-4} 
\multicolumn{1}{|c|}{\multirow{-2}{*}{\textbf{Real}}} & Valid & 41 & 7505 \\
\hline
\end{tabular}
}
\end{minipage}
\begin{minipage}[b]{0.45\linewidth}
Education \\ \\
\centering
\scalebox{0.85}{
\begin{tabular}{cc|c|c|}
\cline{3-4} & & \multicolumn{2}{c|}{\textbf{Prediction}} \\
\cline{3-4} & & Error & Valid \\
\hline
\multicolumn{1}{|c|}{} & Error & 624 & 594 \\
\cline{2-4} 
\multicolumn{1}{|c|}{\multirow{-2}{*}{\textbf{Real}}} & Valid & 35 & 7489 \\
\hline
\end{tabular}
}
\end{minipage}\par
\vspace{0.25cm}
\begin{minipage}[b]{0.45\linewidth}
Government \\ \\
\centering
\scalebox{0.85}{
\begin{tabular}{cc|c|c|}
\cline{3-4} & & \multicolumn{2}{c|}{\textbf{Prediction}} \\
\cline{3-4} & & Error & Valid \\
\hline
\multicolumn{1}{|c|}{} & Error & 694 & 709 \\
\cline{2-4} 
\multicolumn{1}{|c|}{\multirow{-2}{*}{\textbf{Real}}} & Valid & 31 & 6654 \\
\hline
\end{tabular}
}
\end{minipage}
\begin{minipage}[b]{0.45\linewidth}
News and Media \\ \\
\centering
\scalebox{0.85}{
\begin{tabular}{cc|c|c|}
\cline{3-4} & & \multicolumn{2}{c|}{\textbf{Prediction}} \\
\cline{3-4} & & Error & Valid \\
\hline
\multicolumn{1}{|c|}{} & Error & 918 & 1006 \\
\cline{2-4} 
\multicolumn{1}{|c|}{\multirow{-2}{*}{\textbf{Real}}} & Valid & 59 & 9591 \\
\hline
\end{tabular}
}
\end{minipage}\par
\vspace{0.25cm}
\begin{minipage}[b]{0.45\linewidth}
Science and Environment \\ \\
\centering
\scalebox{0.85}{
\begin{tabular}{cc|c|c|}
\cline{3-4} & & \multicolumn{2}{c|}{\textbf{Prediction}} \\
\cline{3-4} & & Error & Valid \\
\hline
\multicolumn{1}{|c|}{} & Error & 736 & 661 \\
\cline{2-4} 
\multicolumn{1}{|c|}{\multirow{-2}{*}{\textbf{Real}}} & Valid & 20 & 7476 \\
\hline
\end{tabular}
}
\end{minipage}
\begin{minipage}[b]{0.45\linewidth}
Overall \\ \\
\centering
\scalebox{0.85}{
\begin{tabular}{cc|c|c|}
\cline{3-4} & & \multicolumn{2}{c|}{\textbf{Prediction}} \\
\cline{3-4} & & Error & Valid \\
\hline
\multicolumn{1}{|c|}{} & Error & 4416 & 3893 \\
\cline{2-4} 
\multicolumn{1}{|c|}{\multirow{-2}{*}{\textbf{Real}}} & Valid & 218 & 46038 \\
\hline
\end{tabular}
}
\end{minipage}
\end{table}

\vspace{0.25cm}

After running the automatic error Web pages detection, the debugging is complete. The composition and final size of our dataset is shown in Table \ref{table:finaldataset}. Combining  Browsing and Searching techniques, we have achieved to collect \textbf{49438} valid Web pages that occupy approx. 17 GB.

\begin{table}[H]
\centering
\caption{Composition and size of the final dataset.}
\label{table:finaldataset}
\begin{tabular}{|c|c|c|c|}
\hline
\textbf{Category}      & 
\begin{tabular}[c]{@{}c@{}}\textbf{Browsing} \\ Webshots\end{tabular} & \begin{tabular}[c]{@{}c@{}}\textbf{Searching}\\ Webshots\end{tabular} & \textbf{Total} \\ \hline
Arts \& Entertainment  & 397 (147 MB)                                                                  & 7355 (2.58 GB)                                                                  & 7752           \\ \hline
Business \& Economy    & 892 (300 MB)                                                                  & 7546 (2.48 GB)                                                                  & 8438           \\ \hline
Education              & 368 (126 MB)                                                                  & 7524 (2.64 GB)                                                                  & 7892           \\ \hline
Government             & 669 (253 MB)                                                                  & 6685 (2.47 GB)                                                                  & 7354           \\ \hline
News \& Media          & 394 (237 MB)                                                                  & 9650 (3.19 GB)                                                                     & 10044            \\ \hline
Science \& Environment & 462 (193 MB)                                                                  & 7496 (2.63 GB)                                                                  & 7958           \\ \hline
\textbf{Total}         & \textbf{3182} (1.22 GB)                                                        & \textbf{46256} (15.99 GB)                                                       & \textbf{49438} \\ \hline
\end{tabular}
\end{table}

\subsection{Publication of the dataset}
\label{subsec:publication}
\begin{figure}[H]
\centering
\caption{\label{fig:osf} Web Pages Dataset Project in OSF.}
 \includegraphics[width=0.5\textwidth]{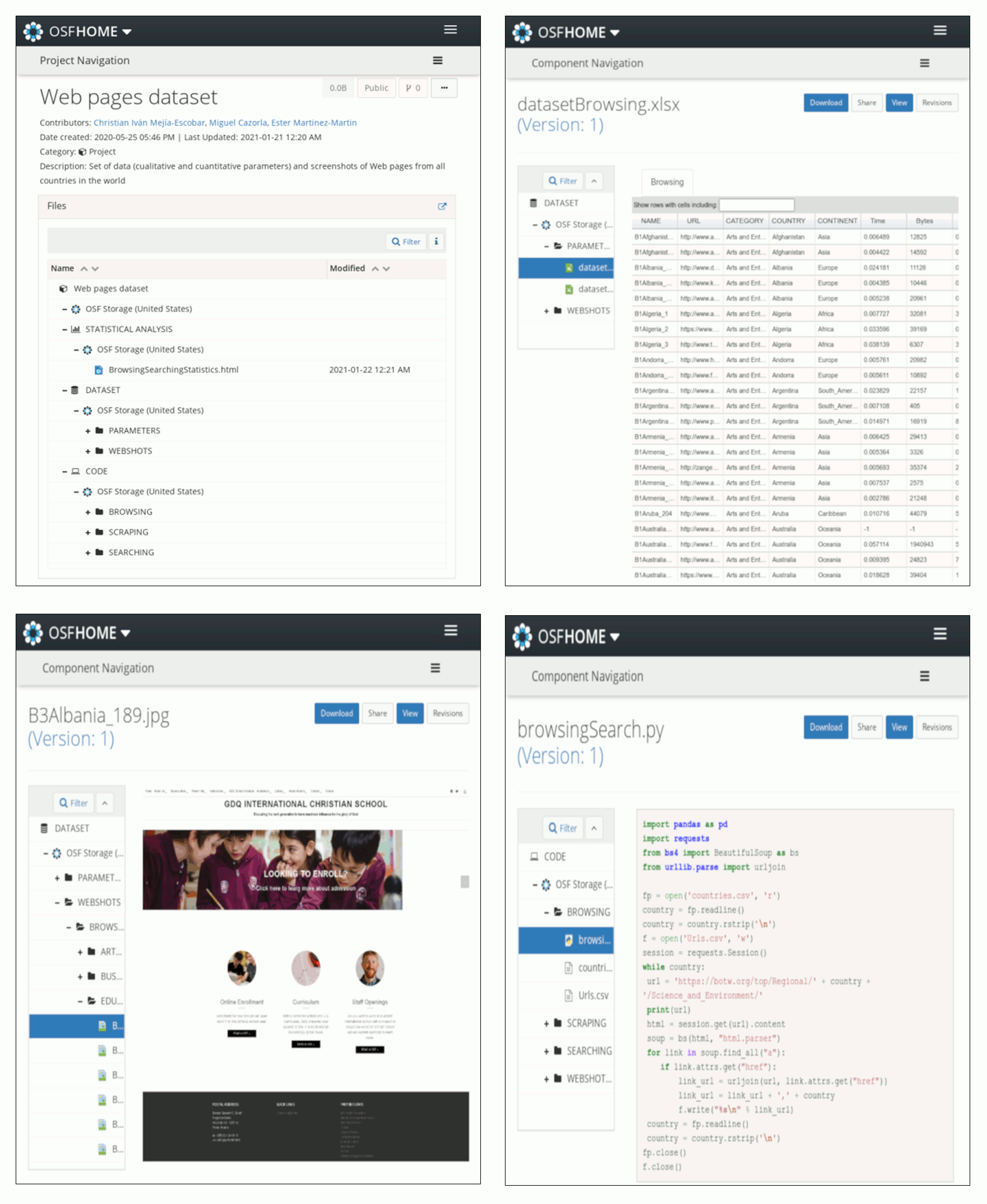} 
\end{figure}

All products generated in this work are public and available through OSF\footnote{\url{https://osf.io/}} (\textit{Open Science Foundation}), a free and open platform to support, disseminate and enable collaboration of scientific research. The OSF Web site offers a user-friendly interface to manage everything related to the project, as shown in Fig.~\ref{fig:osf}.

 The hierarchical organization of the project consists of: a) the dataset divided in the visual part (images in categories), and the textual and numerical part (datasheets); b) the code developed; and c) support documentation. The project is public\footnote{\url{https://osf.io/7ghd2/}} and their resources can be selected and viewed within the same page and downloaded through the button on the top bar.

\section{Statistical analysis}
\label{sec:analysis}
Both sources of the Web pages: Browsing and Searching, are compared through a statistical study supported by the R programming language. Before, we should note that the calculation of statistical indicators and creation of graphs excluded \textit{outliers} due to the high heterogeneity of the variables included in Table~\ref{table:variables}. These values are far from those considered common and may cause distortions in mathematical and visual analysis. By using the well-known rule "1.5 times the \textit{Interquartile Range}", outliers can be identified and omitted. For this reason, the number of values of each of the variables may differ.

\subsection{Qualitative parameters}
\label{subsec:qualitative}
\begin{figure}[H]
\centering
\subfloat[]{\includegraphics[width=0.47\linewidth]{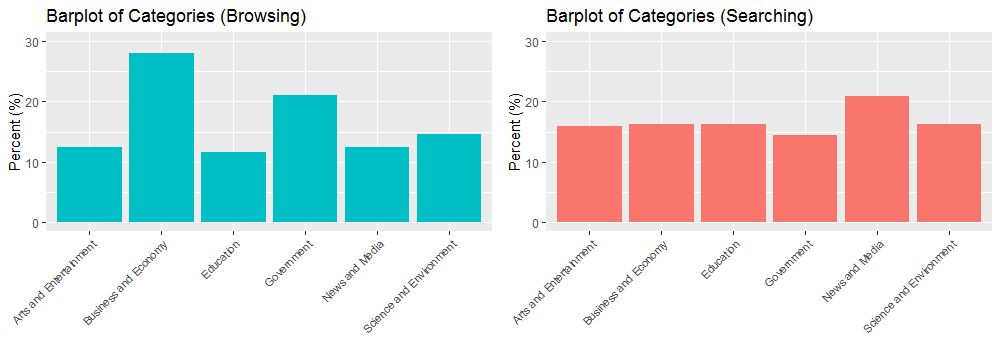}
\label{fig:categories}}
\hfil
\subfloat[]{\includegraphics[width=0.47\linewidth]{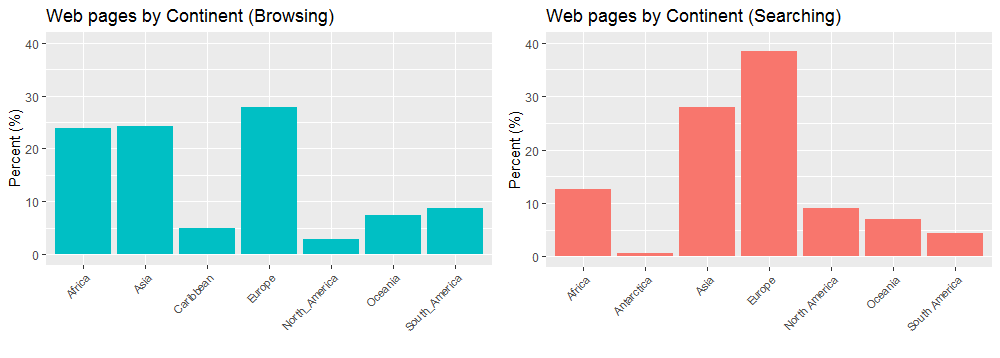}
\label{fig:continents}}

\caption{Distribution of the qualitative parameters about Web pages: a) category and b) continent.}
\label{fig:qualitativeplots}
\end{figure}

Although we tried to obtain an uniformly distributed set of URLs with respect to the categories, the errors cited in the debugging section, caused the results shown in Fig.~\ref{fig:categories}. In Searching, there is less imbalance, in contrast to Browsing, where Web pages related to business, economy and government predominate, possibly due to a greater need for dissemination and economic capacity to register their Web pages in a paid service.

The geographical location of the Web pages is mostly in Europe and Asia, both for Browsing and Searching. Such continents agglomerate a larger number of countries. Moreover, the economic potential could explain why they are at the top. (Fig.~\ref{fig:continents}).

\subsection{Quantitative parameters}
\label{subsec:quantitative}

\begin{figure}[H]
\centering
\subfloat[]{\includegraphics[width=0.47\linewidth]{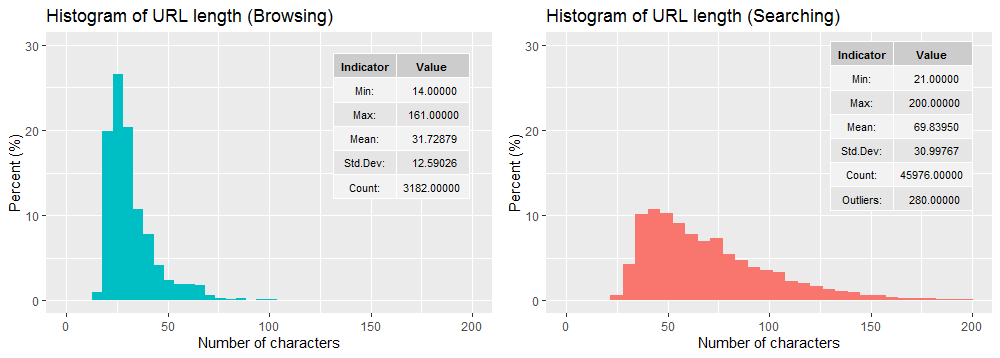}
\label{fig:length}}
\hfil
\subfloat[]{\includegraphics[width=0.47\linewidth]{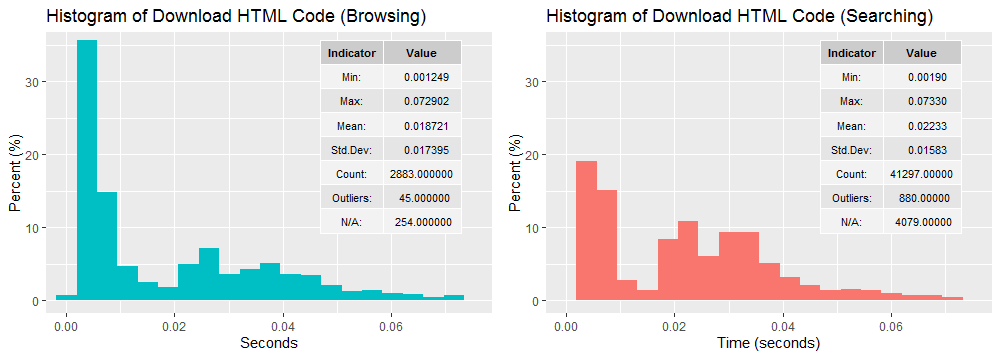}
\label{fig:time}}
\\
\subfloat[]{\includegraphics[width=0.47\linewidth]{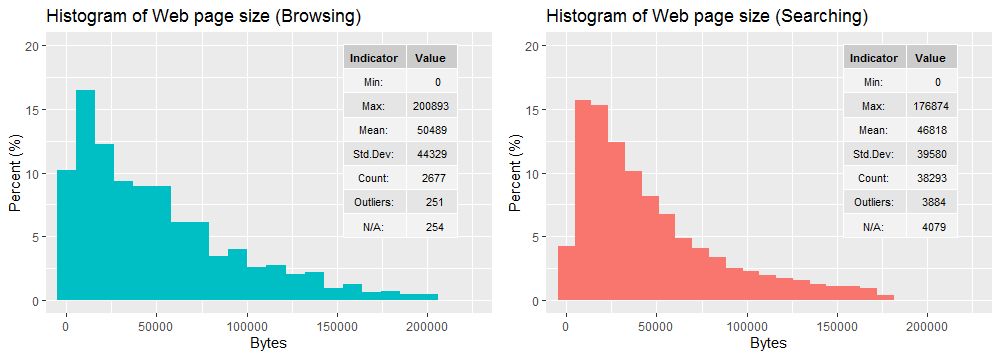}
\label{fig:size}}
\hfil
\subfloat[]{\includegraphics[width=0.47\linewidth]{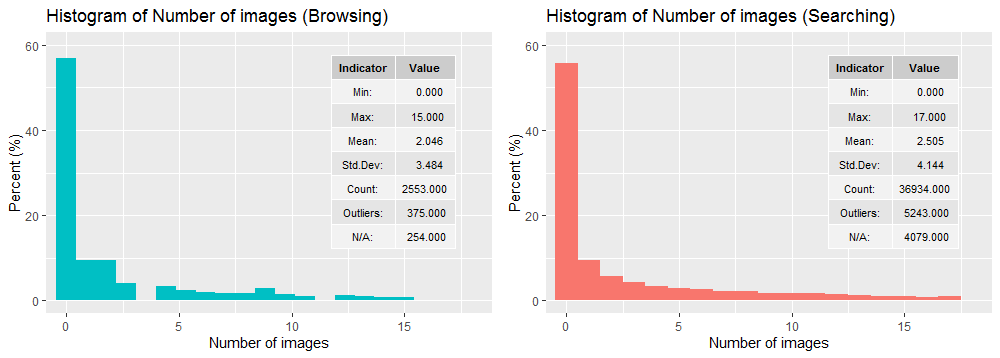}
\label{fig:images}}
\\
\subfloat[]{\includegraphics[width=0.47\linewidth]{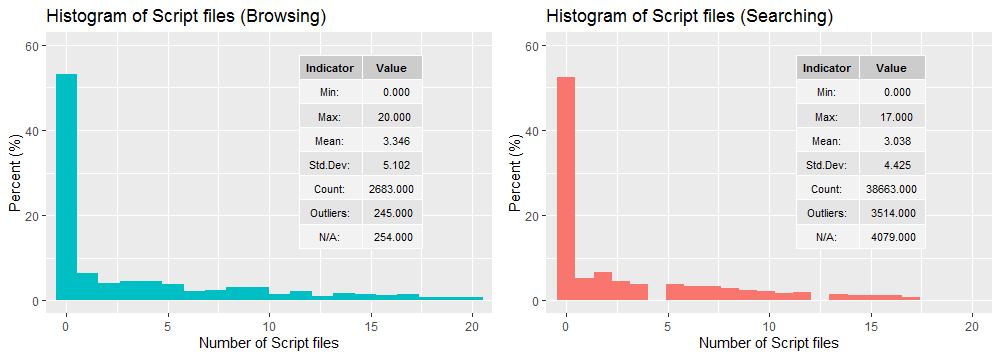}
\label{fig:scripts}}
\hfil
\subfloat[]{\includegraphics[width=0.47\linewidth]{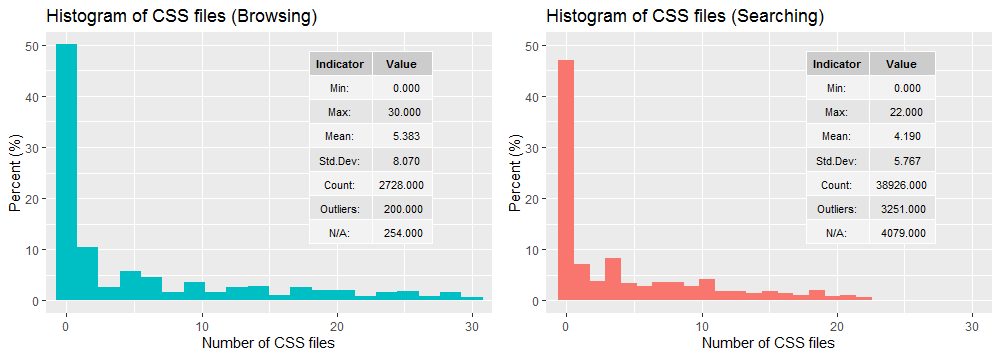}
\label{fig:css}}
\\
\subfloat[]{\includegraphics[width=0.47\linewidth]{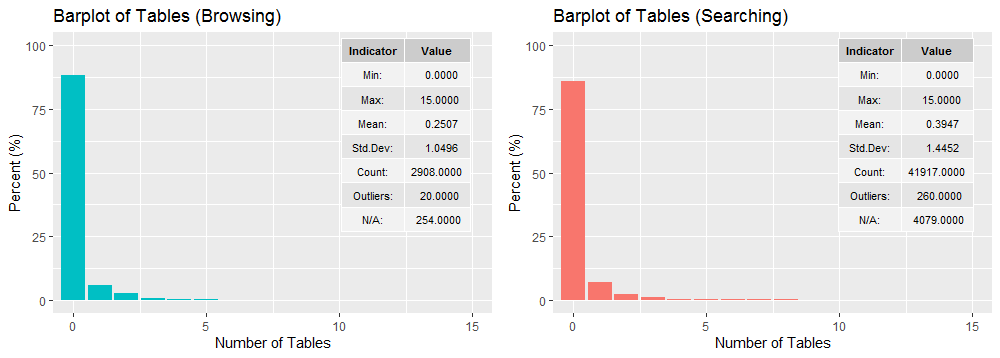}
\label{fig:tables}}
\hfil
\subfloat[]{\includegraphics[width=0.47\linewidth]{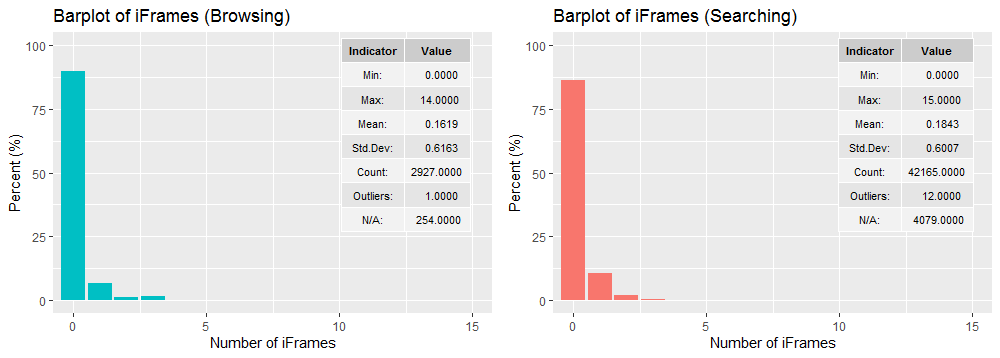}
\label{fig:iframes}}
\\
\subfloat[]{\includegraphics[width=0.47\linewidth]{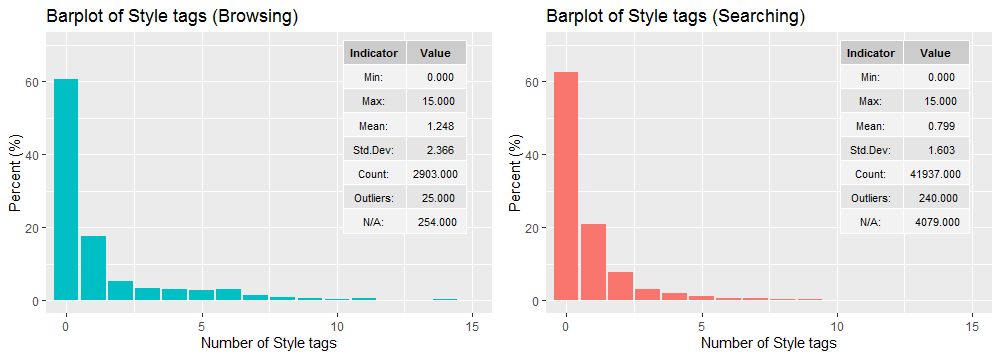}
\label{fig:style}}
\hfil
\subfloat[]{\includegraphics[width=0.47\linewidth]{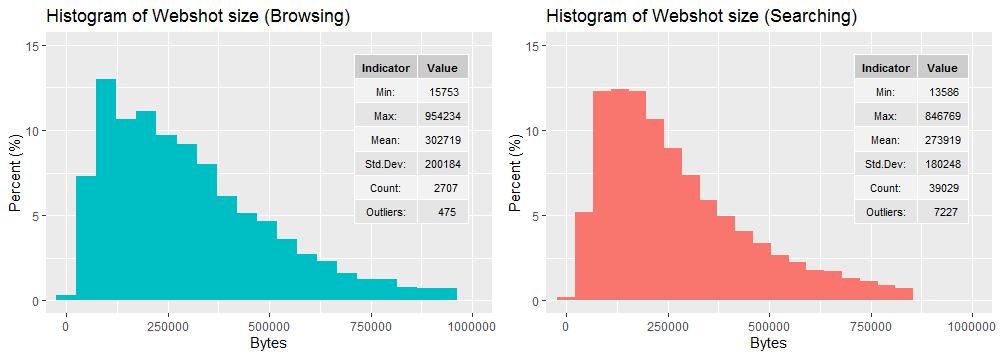}
\label{fig:webshotsize}}
\\
\subfloat[]{\includegraphics[width=0.47\linewidth]{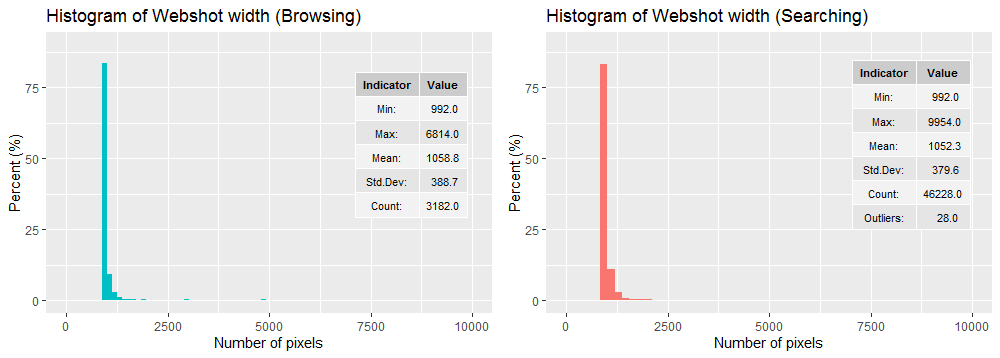}
\label{fig:webshotwidth}}
\hfil
\subfloat[]{\includegraphics[width=0.47\linewidth]{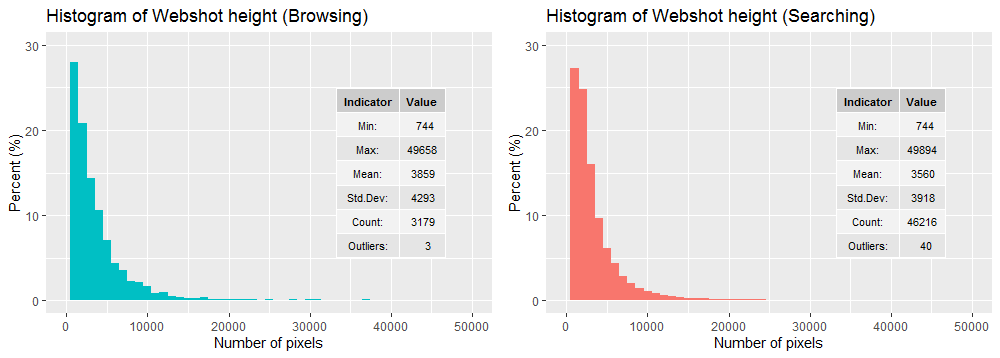}
\label{fig:webshotheight}}

\caption{Distribution of the quantitative parameters about Web pages}
\label{fig:quantitativeplots}
\end{figure}

The variability of Searching is more evident for the number of characters in a URL (Fig.~\ref{fig:length}), since a wider range (the difference between the minimum and maximum) and a higher mean and standard deviation. In both graphs, the values accumulate more at the bottom of the variable and are less frequent at the top, so there is a tail to the right. This behavior is desirable when reading or typing a URL in a browser, and is an indicator that there are not too many levels to reach a particular Web page.

The download time of the source code of the Web pages shown in Fig.~\ref{fig:time} has a similar behavior for Browsing and Searching. In both cases, although there is considerable variability due to the width of the range and standard deviation, the values are around 20 milliseconds and mostly low, which is a benefit for the user who wants to view the Web page in the shortest time possible. According to the errors cited in the debugging section, 254 of 3182 URLs belonging to Browsing were not available, i.e. 7.98\%; while in Searching, 4079 of 46256, i.e. 8.82\%, were not accessible to download the source code, and hence the extraction of the quantitative parameters was not possible.

The behavior of the size in bytes (Fig.~\ref{fig:size}) is almost identical between Browsing and Searching, where the average of the Web pages is approximately 50 KB. Both the variability and the tails of the distributions are practically the same, being favorable for a quick view of the Web page, which is in direct relation to a small size. Although there are still Web pages with a considerable size, which may be due to graphic elements or external objects linked to page.

In Fig.~\ref{fig:images}, close to 60\% of Web pages do not include images within their source code. It seems that the pages present textual information exclusively. However, they might include images through CSS style files, which is a good practice \cite{King}. Although the range of number of images is wide, the average is low, 2 or 3 images per Web page, so there is a tendency to use a few images within a Web page in order to make it lighter.

Non-scripted Web pages exceed 50\% in both cases. Scripts are add-on programs that provide additional functions to Web pages. However, their use may cause incompatibilities with browsers and make the page more complex and heavy. In Fig.~\ref{fig:scripts}, the trend is to minimize the presence of scripts, 3 scripts per Web page on average.

Fig.~\ref{fig:css} shows that approximately 50\% of the Web pages do not use \textit{Cascading Style Sheets} (CSS) files, whose use is recommended as a good practice in Web design. The ideal amount would be one style file per Web page. The average for Browsing and Searching is 5 and 4, respectively, not too far away. The tendency is towards low values, but there are some cases with many style files, being prejudicial to the agile display of the Web page.

The graphs in Fig.~\ref{fig:tables} have a very similar aspect. The majority of Web pages (about 85\%) no longer use tables within the source code. Tables were generally used to structure the content; however, this practice has been replaced with the "div" tag, achieving a more elegant and professional design.

Over 85\% of Web pages do not use iFrames. For Browsing and Searching, the bars in the graph are grouped on the left (Fig.~\ref{fig:iframes}). We can deduce that the embedding of another document in the current HTML document through the "iFrame" tag is disappearing, as there are now better options.

More than half of the Web pages (close to 60\%) no longer use "style" tags in their source code. Both graphs in Fig.~\ref{fig:style} have bars that decrease towards the right. The trend is to minimize the number of such tags, as it is more appropriate to use CSS files. Thus, the source code of a Web page does not extend too much.

The webshot size is decisive in determining how much space our dataset will consume on a storage device. In Fig. \ref{fig:webshotsize}, the size fluctuates over a wide range of values, with an average of about 300 KB. Most of the values are concentrated in low sizes, but with a considerable presence of images with medium and high size. This behavior would require not only a good amount of space but a pre-processing of the images for Machine Learning and Deep Learning applications.

The Searching and Browsing graphs are quite similar for the webshot width (Fig.~\ref{fig:webshotwidth}). In both cases, the first bar, where the minimum is, significantly predominates as the default screenshot sets a width of 992 pixels. The average value is very close to the minimum since almost all images were captured with this default value (about 85\%); however, there are also images with a wider width, especially in Searching with a maximum of almost 10000 pixels.

In the case of the height variable (Fig.~\ref{fig:webshotheight}), the minimum value prevails too, although to a lesser degree (about 28\%), which coincides with the default value set by the screenshot, i.e. 744 pixels. Unlike the previous case, there is a less unbalanced distribution of values, with a wider variability in which the highest accumulation occurs up to 5000 pixels, a considerable accumulation between 5000 and 10000 pixels, and finally, images with a height of up to almost 50000 pixels have been obtained. Considering the width and height, most of the Web pages have a vertical layout. These parameters are closely related to the resolution or quality of the image. The more pixels, the more resolution and quality the image has, but it demands more storage space.

\begin{table}[H]
\centering
\caption{Summary of statistical indicators for quantitative parameters. \label{table:indicators}}
\scalebox{0.95}{
\begin{tabular}{|c|c|c|c|c|c|c|c|c|}
\hline
\multirow{2}{*}{\textbf{Parameter}} & \multicolumn{4}{c|}{\textbf{Browsing}}                                       & \multicolumn{4}{c|}{\textbf{Searching}}                              \\ \cline{2-9} 
                                    & \textbf{Min.}   & \textbf{Max.}         & \textbf{Mean} & \textbf{Std. Dev.} & \textbf{Min.} & \textbf{Max.}   & \textbf{Mean} & \textbf{Std. Dev.} \\ \hline
URL length                          & 14              & 161                   & 31.73         & 12.59              & 21            & 200             & 69.84         & 30.99              \\ \hline
Time (ms)                  & 1.25            & 72.9                  & 18.72         & 17.39              & 1.9           & 73.3            & 22.33         & 15.83              \\ \hline
Size (KB)                           & 0               & 200.89                & 50.49         & 44.33              & 0             & 176.87          & 46.82         & 39.58              \\ \hline
Images                              & 0               & 15                    & 2.05          & 3.48               & 0             & 17              & 2.51          & 4.14               \\ \hline
Scripts                             & 0               & 20                    & 3.35          & 5.1                & 0             & 17              & 3.04          & 4.43               \\ \hline
CSS files                           & 0               & 30                    & 5.38          & 8.07               & 0             & 22              & 4.19          & 5.77               \\ \hline
Tables                              & 0               & 15                    & 0.25          & 1.05               & 0             & 15              & 0.39          & 1.45               \\ \hline
iFrames                             & 0               & 14                    & 0.16          & 0.62               & 0             & 15              & 0.18          & 0.6                \\ \hline
Style tags                          & 0               & 15                    & 1.25          & 2.37               & 0             & 15              & 0.8           & 1.6                \\ \hline
Size (KB)                   & 15.75           & 954.23                & 302.72        & 200.18             & 13.59         & 846.77          & 273.92        & 180.25             \\ \hline
Width (px)                  & 992             & 6814                  & 1058.8        & 388.7              & 992           & 9954            & 1052.3        & 379.6              \\ \hline
Height (px)                 & 744             & 49658                 & 3859          & 4293               & 744           & 49894           & 3560          & 3918               \\ \hline
\end{tabular}
}
\end{table}

Finally, Table \ref{table:indicators} summarizes the main statistical indicators for the quantitative parameters of the Web pages, both for Browsing and Searching set.

\section{Case study: Multi-class categorization of Web pages}
\label{sec:caseuse}
The Web is a global communication platform where the volume of information available is enormous, grows rapidly, and covers any topic. Search and recovery service providers are constantly working on improving mechanisms to manage this information effectively. Classification is a basic technique to deal with this problem. Since doing it manually is not practical, the automatic classification of Web pages is the recommended method and has motivated several research and development works.

The classification of Web pages, also called \textit{Web categorization}, determines whether a Web page or Web site belongs to a category in particular. For example, judging whether a page is about "arts", "business", or "sports" is an instance of subject classification \cite{Qi}. This is usually done by analyzing both the textual content and underlying HTML code. However, the visual appearance is also an important part of a Web page, and many topics have a distinctive visual appearance, e.g., Web design blogs have a highly designed visual appearance, whereas newspaper sites will have a lot of text and images \cite{DeBoer}.

Here, we present an automatic categorization of Web pages according to topic or subject and based exclusively on their visual appearance. We take advantage of the dataset generated in this work, formed by images (webshots) belonging to 6 categories: arts and entertainment, business and economy, education, government, news and media, and science and environment. Therefore, the problem becomes a multi-class categorization.

We implemented a model of \textit{Deep Learning} with a \textit{Convolutional Neural Network (CNN)}. In essence, a learning process with the webshots collected in order to achieve an acceptable accuracy and then make predictions. We hope to capture features (difficult to identify manually) that can distinguish categories, predict to which of them a Web page would belong, analyze the difficulty of the topic classification of the Web pages and verify if there are particular patterns for each category.

It is important to highlight that the following results have been selected from a series of various experiments, where different models and architectures were tested using the whole dataset and parts of it. The developed code, as well as the weights of the adjusted deep learning model, are publicly accessible\footnote{\url{https://osf.io/8zfh2}}.

The best results were obtained with the \textit{Transfer Learning} technique and the images of the Browsing dataset. This may be since in a Web directory such as \textit{BOTW}, the Web pages go through a rigorous registration process under the supervision of human specialists, so the Web pages have a better distinction and categorization.

The dataset for the multi-class categorization is constituted only by the webshots of the set of Browsing, which has been organized and split according to Table \ref{table:CategorizationDataset}. For the training process, we have a balanced amount of data, i.e. the same number of images for each category, to avoid possible preferences. The category with fewer images (Education) was the basis for randomly selecting the same number of images in the other categories. One image of 3.68 MB and resolution of 992x30154 was discarded because the Python imaging library does not open larger images to avoid malicious attacks. So, each category has 367 images, a total of 2202 images, 80\% for training, whereas 20\% remaining is destined for validation (both sets randomly selected).

\begin{table}[H]
\centering
\caption{Dataset split for multi-class categorization.}
\label{table:CategorizationDataset}
\scalebox{0.75}{
\begin{tabular}{|c|c|c|c|c|}
\hline
\textbf{Category} & \textbf{Webshots} & \textbf{Dataset} & \textbf{Train (80\%)} & \textbf{Val. (20\%)} \\
\hline
Arts \& Entertainment & 397 & 367 & 293 & 74 \\
\hline
Business \& Economy & 892 & 367 & 293 & 74 \\
\hline
Education & 368 & 367 & 293 & 74 \\
\hline
Government & 669 & 367 & 293 & 74 \\
\hline
News \& Media & 394 & 367 & 293 & 74 \\
\hline
Science \& Environment & 462 & 367 & 293 & 74 \\
\hline
\textbf{TOTAL} & \textbf{3182} & \textbf{2202} & \textbf{1758} & \textbf{444} \\
\hline
\end{tabular}
}
\end{table}

\begin{figure}[H]
\centering
    \includegraphics[width=0.25\textwidth]{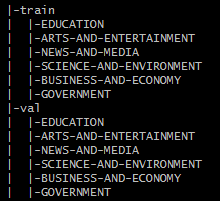}
    \caption{\label{fig:DatasetCategorizationStructure} Directory structure by categories for training and validation.}
\end{figure}

The images are stored within the directory structure displayed in Fig.~\ref{fig:DatasetCategorizationStructure}. Within the main folder of the dataset the division in training and validation, and the sub-folders represent the categories, which have been named as the topics considered in this work.

After organizing the images, a pre-processing step is convenient to normalize the image's pixel values (integers between 0 and 255) to the scale of values between 0 and 1; and resize to 224x224 pixels recommended for the model, because the images in the dataset have different dimensions (width and height). Both are a common practice that helps speed up the process of training.

In the training phase, several models were tested with a variety of options to achieve greater accuracy. The final model exploits the \textit{Transfer Learning} technique using \textit{ResNet} \cite{Resnet}, a competitive CNN pre-trained on the ImageNet dataset (above 14 million images belonging to 1000 categories) and winner of ImageNet challenge in 2015.

\begin{figure}[H]
\centering
    \includegraphics[width=0.95\textwidth]{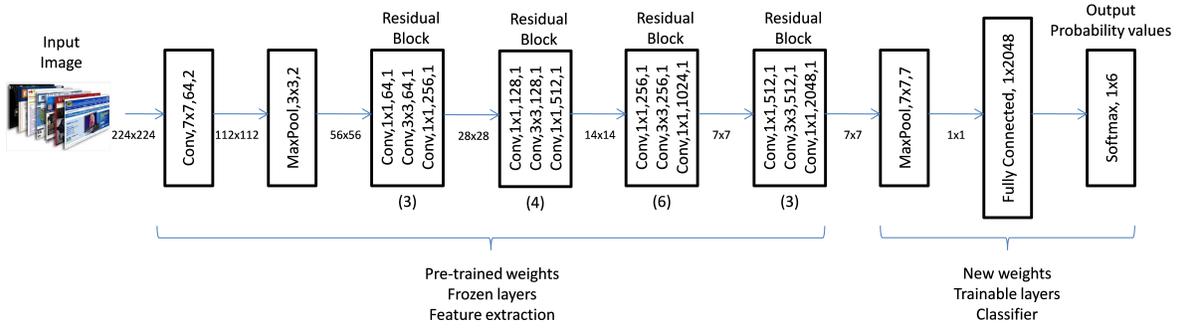}
    \caption{\label{fig:resnet50} Model's architecture based on ResNet-50.}
\end{figure}

Despite more up-to-date models, ResNet is still very popular for Transfer Learning implementations. We used \textit{ResNet-50} that is 50 layers deep, whose convolutional basis is kept for feature extraction, while the classifier part is replaced by a new one that will predict the probabilities for 6 classes corresponding to our categories (Fig.~\ref{fig:resnet50}).

\begin{figure}[H]
\centering
    \includegraphics[width=0.5\textwidth]{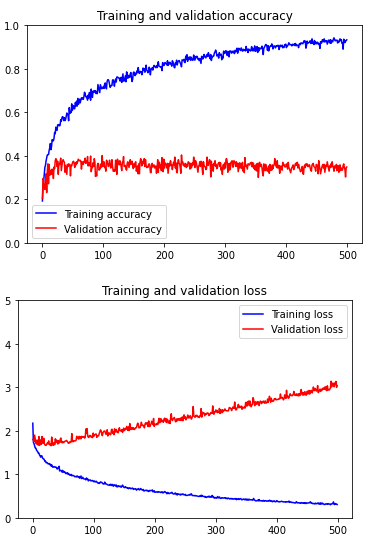}
    \caption{\label{fig:categorizationaccuracy} Accuracy and loss in training and validation phases.}
\end{figure}

Only the classifier layers are trained on our dataset. After 500 iterations (epochs) of the whole training set (1758 images) in groups of 32 images (batch\_size), an accuracy of \textbf{94.26\%} was obtained and \textbf{40.38\%} in the validation phase. The evolution of the process is summarized in the following training and validation curve graphs (Fig.~\ref{fig:categorizationaccuracy}).

A suitable solution to this problem must meet: a) high training accuracy; b) validation and training curves very close to each other; and c) small difference between the validation and training error. According to the graphs, only the first item is accomplished, so the model learned very well but does not for generalization, i.e. to classify new images acceptably. Although we increased the data, tuned the hyperparameters, and applied regularization techniques such as dropout, neither accuracy is improved nor overfitting is significantly reduced.

\begin{figure}[H]
\centering
    \includegraphics[width=0.4\textwidth]{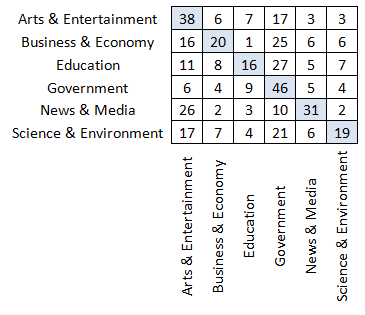}
    \caption{\label{fig:categorizationmatrix} Confusion matrix for validation data.}
\end{figure}

For a better understanding of the results, Fig.~\ref{fig:categorizationmatrix} displays the confusion matrix with the validation data.

The model is correct in most cases if we focus on the categories of arts and entertainment, government, and news and media; however, the number of successes is low. This test takes the validation data, a total of 444 images, achieving an accuracy of \textbf{38.29\%}, according to the confusion matrix. For remaining categories, the model gets significantly confused. The classification of these categories is a very hard problem. Nowadays, the composition of Web pages is becoming more complex, and the content has a high variability of visual features, even within the same category.

\section{Conclusions}
\label{sec:conclusions}
We combined different types of data (text, numbers, and images) in a single and extensive dataset about Web pages. Although the data is organized separately, the qualitative and quantitative parameters are linked to the webshots through an alphanumeric identifier whose name allows us to know the Web page's source, category, and country of origin.

We were able to collect 3609 webshots from Browsing (Table \ref{table:DatasetSplit}) and 54565 from Searching (Table \ref{table:classificationresults}), a total of 58174 webshots; however, the final dataset was reduced to 49438 due to the elimination of Web pages with error messages. This type of non-valid pages is equivalent to 15\%, a significant value that reflects a problem on the Internet that affects webmasters, search engines, and users in general. To address this problem, more efficient debugging processes are needed. As an approach, we implemented an automatic detection of error Web pages based on a CNN model from scratch achieving acceptable accuracy.

The methodology described for the generation of the Web pages dataset could be adapted without inconvenience in several problems that involve the collection, organization, analysis, and publication of large amounts of data. Much of the workflow was automatized using Python and R scripts that allowed us to collect URLs and their webshots. By using scraping, several parameters have been extracted from each Web page. In this way, thousands of Web pages, indexed by Google and BOTW Web Directory, organized by categories (Arts and Entertainment, Business and Economy, Education, Government, News and Media, Science and Environment), and geographical location (country and continent), are available for public use.

The statistical analysis of the quantitative parameters: download time, size, number of images, scripts, CSS files, number of tables, iFrames, and style tags, show similar behavior. These variables have a very heterogeneous distribution, high variability, and tend to be strongly concentrated in the low values (Table \ref{table:indicators}). This suggests that Web design follows an implicit rule of optimization of all these parameters, since the higher their values, the longer the download and display time of the page would increase, causing the consequent user's discomfort.

Regarding the multi-class CNN model, the results showed that the automatic categorization of Web pages based exclusively on visual appearance (webshot) is a highly complex problem. Our dataset has proven to be difficult to classify; the difficulty increases when the categories cover a wide range of topics (like the case presented here). In addition, within each topic there is also a lot of variability in the visual aspect of the images.

Although it is not possible to reliably distinguish between the categories in the dataset using only the webshots, the successes of the Deep Learning model for Web categorization, especially for government and, arts and entertainment categories, allow us to presume that it is feasible to identify distinctive visual patterns, which may be part of a next work. A high level of accuracy has not been achieved; however, it can serve as a baseline for future research. We suppose that increasing the dataset, preprocessing the images to have the same size and resolution, cropping and scaling the Web pages, could improve the results.

\section{Future work}
\label{sec:future}
Our work can motivate the following developments:
\begin{itemize}
    \item Extending the dataset provided, increasing the categories, URLs, webshots, and other qualitative and quantitative parameters.
    \item Achieving a higher level of automation in the process of collecting, organizing, and capturing webshots.
    \item Considering alternative URLs sources, that is, a search engine other than Google for Searching, and a Web directory other than BOTW for Browsing.
    \item Improving the accuracy achieved in multi-class categorization of Web pages with deeper and more recent convolutional neural networks.
\end{itemize}

\bibliographystyle{unsrtnat}

\end{document}